\documentclass[10pt,twocolumn,letterpaper]{article}

\usepackage{lipsum}
\usepackage{iccv}
\usepackage{times}
\usepackage{epsfig}
\usepackage{graphicx}
\usepackage{amsmath}
\usepackage{amssymb}
\usepackage{booktabs}
\usepackage{multirow}
\usepackage{color}
\newtheorem{theorem}{Theorem}[section]

\newcommand\blfootnote[1]{%
\begingroup
\renewcommand\thefootnote{}\footnote{#1}%
\addtocounter{footnote}{-1}%
\endgroup
}

\usepackage[pagebackref=true,breaklinks=true,letterpaper=true,colorlinks,bookmarks=false]{hyperref}

\iccvfinalcopy 

\def\iccvPaperID{4505} 

\ificcvfinal\fi

\begin{document}

\title{ESSAformer: Efficient Transformer for Hyperspectral Image Super-resolution}

\author{Mingjin Zhang$^1$, Chi Zhang$^{1*}$, Qiming Zhang$^{2*}$, Jie Guo$^1$, Xinbo Gao$^3$, Jing Zhang$^2$ \\
$^1$ Xidian University, China\\
$^2$ The University of Sydney, Australia\\
$^3$ Chongqing University of Posts and Telecommunications, China\\
{\tt\small mjinzhang@xidian.edu.cn, ch\_zhang@stu.xidian.edu.cn, qzha2506@uni.sydney.edu.au, } \\ 
{\tt\small jguo@mail.xidian.edu.cn, gaoxb@cqupt.edu.cn, jing.zhang1@sydney.edu.au}
}

\maketitle
\ificcvfinal\thispagestyle{empty}\fi

\begin{abstract}

    Single hyperspectral image super-resolution (single-HSI-SR) aims to restore a high-resolution hyperspectral image from a low-resolution observation. However, the prevailing CNN-based approaches have shown limitations in building long-range dependencies and capturing interaction information between spectral features. This results in inadequate utilization of spectral information and artifacts after upsampling. To address this issue, we propose ESSAformer, an ESSA attention-embedded Transformer network for single-HSI-SR with an iterative refining structure. Specifically, we first introduce a robust and spectral-friendly similarity metric, \ie, the spectral correlation coefficient of the spectrum (SCC), to replace the original attention matrix and incorporates inductive biases into the model to facilitate training. Built upon it, we further utilize the kernelizable attention technique with theoretical support to form a novel efficient SCC-kernel-based self-attention (ESSA) and reduce attention computation to linear complexity. ESSA enlarges the receptive field for features after upsampling without bringing much computation and allows the model to effectively utilize spatial-spectral information from different scales, resulting in the generation of more natural high-resolution images. Without the need for pretraining on large-scale datasets, our experiments demonstrate ESSA's effectiveness in both visual quality and quantitative results. The code will be released at \href{https://github.com/Rexzhan/ESSAformer/tree/main}{ESSAformer}.

\end{abstract}

\blfootnote{*Corresponding author}

\section{Introduction}
Hyperspectral imaging (HSI) involves densely sampling spectral features with many narrow bands to encode rich spectral and spatial structure information for material differentiation. It has been widely used in various applications. Hyperspectral image super-resolution (HSI-SR) aims to generate high-resolution HSI from low-resolution HSI and can be categorized into two approaches: single-HSI-SR \cite{li2017hyperspectral,mei2017hyperspectral,jiang2020learning,zhang2021difference} and pansharpening \cite{loncan2015hyperspectral,zheng2020hyperspectral,palsson2017multispectral,li2018fusing}. This paper focuses on the challenging single-HSI-SR task, which aims to restore high-resolution HSI from a single low-resolution HSI without auxiliary images.

Conventional methods for single-HSI-SR involve designing a mapping function between low-resolution and high-resolution HSI using hand-crafted priors such as low-rank approximation and sparse coding \cite{irmak2018map,wang2017hyperspectral,huang2014super,fu2017adaptive}. However, with the fast development of deep learning, powerful convolutional neural networks (CNNs) have led to significant progress in the single-HSI-SR task \cite{li2017hyperspectral,li2018single,mei2017hyperspectral,yang2019multi,li2020mixed,wang2020spatial,jiang2020learning}. These CNN-based approaches usually use deep neural networks to formulate and learn the mapping function in an end-to-end manner using abundant training data pairs. As a result, they achieve significant improvements in both visual quality and quantitative metrics.

However, the CNNs methods show limitations in solving the single-HSI-SR task. There exists a significant amount of long-range information in high-dimensional data of HSI, while the most prevailing CNNs focus on local features captured by the convolutional kernels \cite{li2017hyperspectral,li2020mixed,wang2020spatial,jiang2020learning,li2018single,mei2017hyperspectral,yang2019multi}. The limited receptive field in the network can thus hinder the models' representation ability. Consequently, unwished artifacts, such as the blocking ones, may appear and affect the model's generation quality. To address this issue, we take an attempt to propose a Transformer model for the single-HSI-SR task. 
The attention mechanism introduced in Vision Transformers allows them to capture long-range dependencies and provide powerful representations, leading to superior performance compared to CNNs in many vision tasks \cite{carion2020end,zhu2020deformable,zheng2021rethinking,chen2021pre,zhou2018end}.

While the long-range dependency advantage of Vision Transformers can potentially address the aforementioned issues, it cannot be directly applied to single-HSI-SR. Firstly, Vision Transformers typically require a large amount of data to learn inductive biases and produce reliable results. However, the difficulty in obtaining HSIs limits the collection of large-scale datasets, which poses a particular challenge compared to the millions of images available in RGB image datasets, thus hindering the training of Vision Transformers. Secondly, while Transformers can handle long-range dependencies, the self-attention process has a quadratic computation complexity of $\mathcal{O}(N^2)$ with respect to the token sequence $N$. This results in a massive computation burden for the network, particularly for ultra-high resolution HSI.

To address the above issues, we propose a novel Transformer model called ESSAformer. The ESSAformer is designed with several adaptations. First, it utilizes an iterative downsampling and upsampling strategy to capture both global and local information at different scales and encode the detailed content of the hyperspectral images. Second, we propose to replace the conventional dot product (cosine similarity) with the robust and spectral-friendly spectral correlation coefficient of the spectrum, called SCC. Compared to traditional cosine similarity, the SCC has desirable properties such as spectral-wise shifting and scaling equivalence. This makes the model insensitive to amplitude-level changes in spectral curves caused by occlusions or shadows. As a result, SCC brings inductive biases into models, facilitates training efficiency, and even enables from-scratch training of Transformer models on small datasets. Third, we propose to formulate the attention as kernelized ones to decrease the computation burden. Technically, we integrate SCC into a nonlinear square exponential kernel, \ie, Mercer's kernel, and then express SCC as a dot product of two individual terms according to the Mercer theorem. Subsequently, we change the multiplication order of self-attention, i.e., multiplying keys and values first and then queries, and thus lower the attention complexity from quadratic $\mathcal{O}(N^2)$ to linear $\mathcal{O}(N)$. Such a pipeline significantly relieves the computation burden since the token number $N$ for high-resolution HSIs is usually significantly long. Consequently, we propose the novel SCC-kernel-based self-attention, called ESSA, and a new ESSAformer Vision Transformer architecture for the single-HSI-SR task.

Thanks to the proposed ESSA, our model efficiently enlarges the receptive field without imposing a significant computation burden, thus allowing the features to attend to the entire feature map at each layer and gather sufficient information. Consequently, ESSA effectively addresses the artifact issues caused by limited and inconsistent receptive fields between any two pixels in hyperspectral images, resulting in more natural high-resolution HSI generation. Unlike other attention variants \cite{zhang2022vsa,xia2022vision}, our ESSA does not bring extra parameters and effectively introduces inductive biases. Consequently, the ESSAformer Transformer model obtains state-of-the-art performance on three public datasets without the need for pretraining.

In summary, this paper makes three main contributions. First, we introduce the use of Vision Transformer for the single-HSI-SR task and propose the ESSAformer model with strong learning ability. Second, we present a novel and efficient SCC-kernel-based self-attention method, called ESSA. The approach significantly reduces the computation and data-hungry issues in the original Vision Transformer and helps the model better fit the single-HSI-SR task. Third, extensive experiments have been conducted to thoroughly analyze the proposed model, and the state-of-the-art performance on three popular datasets demonstrates its superiority regarding both visual quality and objective metrics.

\begin{figure*}[t]
  \centering
  
    \includegraphics[width=\linewidth]{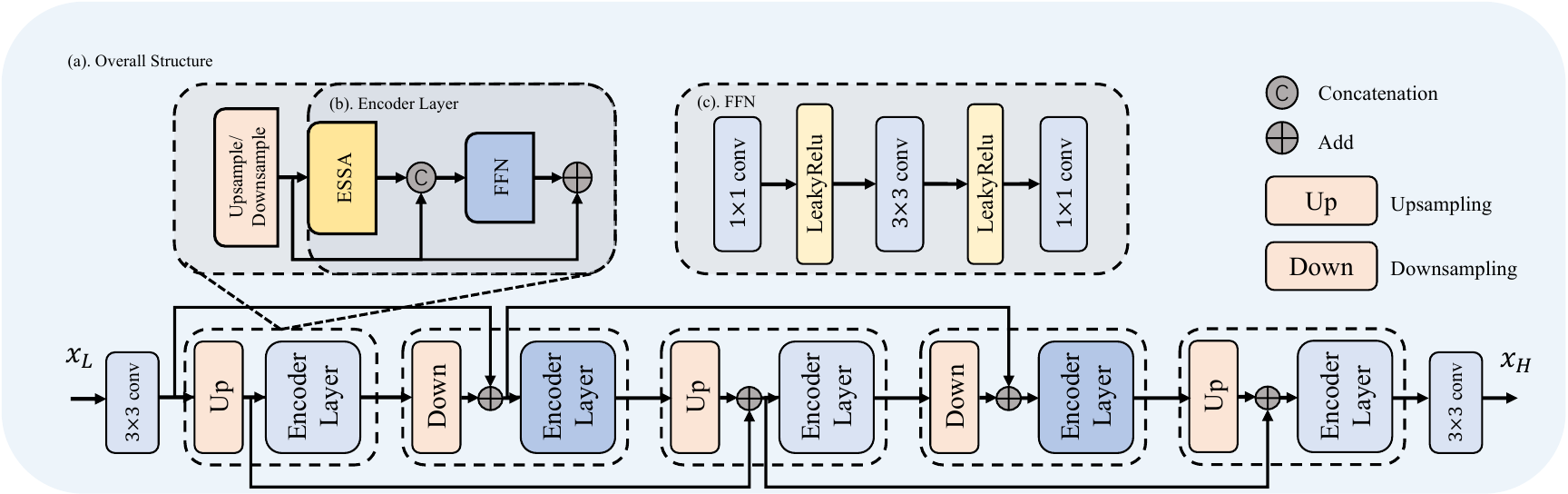}

   \caption{Overview of the proposed ESSAformer. The model is constructed by stacking upsampling/downsampling and encoder layers. The pipeline follows an iterative refinement process to handle the feature representations at different scales, thereby better encoding details and contextual information. All encoder layers having the same input resolution, \ie, the same color, share weight for a lightweight design. }
   \label{fig:structure}
\end{figure*}

\section{Related Work}
\label{sec:related works}

\subsection{HSI super-resolution}
For HSI-SR, the existing prevailing approaches are mostly based on CNNs. Since the researchers first apply CNN \cite{li2017hyperspectral} into this task by proposing a deep spectral difference network, the architecture design for HSI-SR has attracted much attention from the community. For example, a 3D full convolutional network (3D-FCNN) \cite{mei2017hyperspectral} is proposed to recover high-quality HSI without any auxiliary information. Besides, a mixed 2D/3D convolutional network (MCNET) \cite{li2020mixed} and a bidirectional 3D convolutional network (Bi-3DQRNN) are proposed to take into account the forward and backward spectral dependence of HSIs \cite{fu2021bidirectional}. Besides the 3D CNNs, the recursive strategy also demonstrates its effectiveness \cite{li2018single}, which utilizes grouped recursive modules in the global residual structure to depict the complicated non-linear mapping function. Such group strategy has inspired various works to avoid over-processing the redundancy in HSIs and save computational costs. For example, SSPSR \cite{jiang2020learning} employs grouped convolutional layers and channel attention to exploit spectral correlation. Besides, Zhang \etal \cite{zhang2021difference} develop a difference curvature network (DCM-NET) in light of the grouping strategy. 
Nevertheless, prevailing methods are not beyond the limitation of CNN. They are not good at capturing long-range dependencies and modeling spatial-spectral correlation, resulting in insufficient utilization of spectral information and thus unwished artifacts after super resolution. To this end, we propose a novel Transformer model ESSAformer to solve super-resolution in HSIs.

\subsection{Efficient vision transformer}
Transformer is originally proposed in the natural language process field \cite{vaswani2017attention,devlin2018bert,brown2020language,wolf2020transformers}. After Dosovitskiy \etal \cite{dosovitskiy2020image} fed images into a pure Transformer and gained great success on various image recognition tasks, Vision Transformer has received much attention from researchers in various computer vision tasks \cite{carion2020end,zhu2020deformable,zheng2021rethinking,chen2021pre,zhou2018end}. Since the quadratic computation complexity of attention hinders the Transformer application, especially for high-resolution images, several works are proposed to relieve such issue. For example, Geng \etal \cite{geng2021attention} leverages matrix decomposition to substitute the original self-attention while modeling the dependence between different tokens. Similarly, self-attention is approximated as a linear dot-product of kernel feature maps \cite{katharopoulos2020transformers} to avoid huge computation in attention. In contrast, kernelized attention \cite{yuan2021tokens,luo2021stable} aims to find kernels to approach the attention matrix and relieve the computation by changing the multiplication order. Our ESSA falls in this track to improve the computation efficiency. However, previous works target the original attention with RGB images, while ESSA proposes an efficient attention method particularly designed for HSIs. Specifically, ESSA fully considers the characteristics of the hyperspectral field and brings channel-wise inductive bias into the models for better restoration performance.

Efficient Vision Transformer has also demonstrated its effectiveness in image restoration tasks. For example, SwinIR \cite{liu2021swin,liang2021swinir} uses window attention to calculate attention within local windows instead of the whole feature to reduce the computation cost. Stoformer \cite{xiaostochastic} studies the window partition mechanism and proposes a stochastic shifting method. Wang \etal \cite{wang2022uformer} designs a locally-enhanced window-based attention mechanism and a U-Shape model architecture for the restoration task, while CAT \cite{zheng2022cross} extends the window shape to rectangles. Different from them, Lee \etal \cite{lee2022knn} established local attention with non-local connectivity using local-sensitive hashing. Besides the spatial-wise attention, channel-wise attention also proves to work well for high-resolution image restoration tasks \cite{zamir2022restormer,cai2022mask}. They have linear complexity and are most related to our methods. However, ESSA still conducts spatial-wise attention. With strict theoretical support, it uses kernelizable techniques to enable the multiplication exchange, which is significantly different from channel-wise attention in motivation, mathematical formulation, and performance.

\section{Method}

In this section, we will first introduce the overall structure of our proposed ESSAformer model. Then ESSA's implementation details, theoretical support, and complexity analysis are presented in the following part.

\subsection{Overall structure}
The overall structure is presented in Figure~\ref{fig:structure}. Given the predefined scaling ratio $s$ and low-resolution HSIs $ \mathbf{x}_L \in R^{h\times w\times c}$, 
ESSAformer outputs the high-resolution HSIs $ \mathbf{x}_H \in R^{sh \times sw \times c}$ through the learned mapping function $M(\cdot)$ of low to high resolution, where $h,w,c$ denote the height, width and channel of HSIs respectively and $s$ usually set to 2, 4, 8, etc.
\begin{equation}
  \mathbf{x}_{H} = M(\mathbf{x}_{L})
  \label{eq:mapping}
\end{equation}
The channel dimension $c$ is usually larger than that in RGB images for HSIs. Each channel describes the real world at discrete bands from a wide range of continuous spectrums.

The model first projects the raw HSIs into features with a projection layer and then contains several stages to sequentially process the features. At the beginning of each stage is a rescaling module to upsample/downsample the feature maps. Then the encoder layer that has an ESSA and an FFN module follows to encode the features, as shown at the top of Figure~\ref{fig:structure}. For each upsampling/downsampling module, multiple projections and PixelShuffle/PixelUnshuffle \cite{shi2016real} layers with a rescale ratio of 2 are used sequentially until they output the feature map to the expected resolution. For example, we use 3 layers in the rescaling module for the required scaling ratio $s=8$. The features channel dimension keeps after both upsampling and downsampling modules. In encoder layers, the feature map after ESSA is concatenated with ESSA's input and then fed into the feed-forward layer (FFN), which is consisted of several convolution and activation layers following the common practice \cite{li2021localvit}.

These sequential stages follow an up-down strategy and thus construct an iterative refinement process by encoding feature representations at different scales, allowing the model to effectively capture content details and contextual information. Each stage includes one encoder layer and all stages that have the same input resolution (the same color in Figure~\ref{fig:structure}) share weights to enable a lightweight model design. Besides, ESSAformer uses a residual connection between features right after the up/downsampling layers of the adjacent three stages as illustrated in Figure~\ref{fig:structure}. This approach enables the model to utilize multi-scale features and generate better feature representations. Finally, after the last stage with an upsampling module, a convolution layer with a $3\times3$ kernel projects the feature maps into the required channel dimension $c$, resulting in high-resolution HSIs $\mathbf{x}_H$.

\begin{figure}[t]
  \centering
  
    \includegraphics[width=\linewidth]{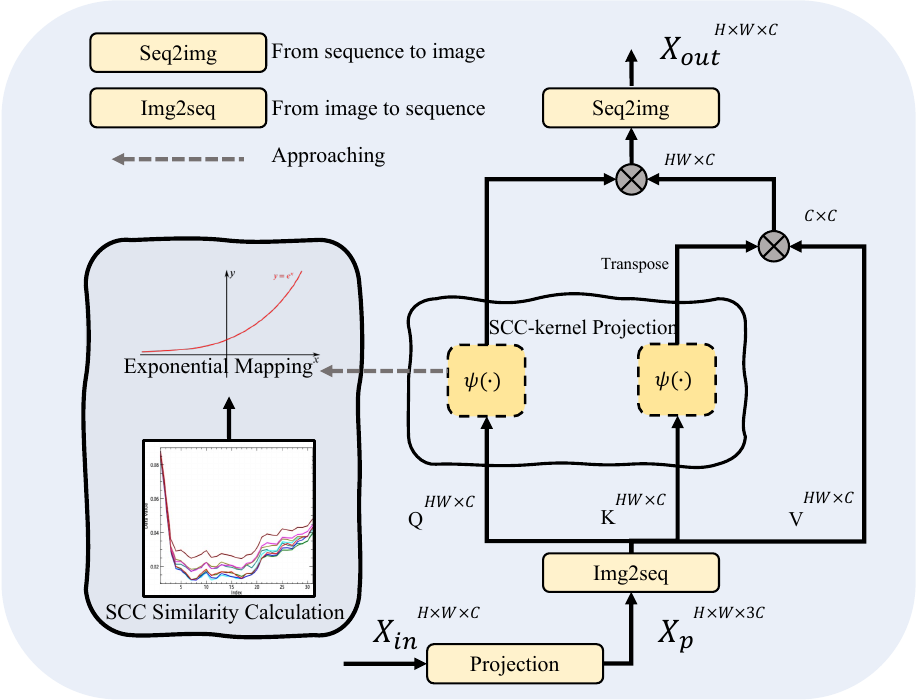}

   \caption{The structure of our ESSA.}
   \label{fig:ESSA}
\end{figure}

\subsection{SCC self-attention}
As one of the cores of Transformer, self-attention enlarges the dependence distance by attending to the feature at each position. We will start from the original attention process and then introduce SCC self-attention that introduces channel-wise inductive biases to improve the data efficiency. We take the input resolution $H\times W$ for example.

Given the input features in $ R^{H\times W\times C}$, a projection layer first embeds them into three token sequences, i.e., $Q, K, V \in R^{N\times C}$, where $N$ equals to $H\times W$, i.e., the sequence length, and $C$ is the channel dimension. The attention function computes the dot products of queries with all keys and applies a softmax function as the weights on the values. We take the single-head self-attention for simplicity and the function is thus given by:

\begin{equation}
  {\rm Attention}(Q,K,V) = {\rm softmax}(\frac{QK^T}{\sqrt{C}})V
  \label{eq:attention}
\end{equation}

Then we will introduce the proposed SCC self-attention. It considers the specific characteristics in hyperspectral images and brings inductive biases to improve data efficiency and representation ability.

To make attention spectral-friendly, we utilize a robust spectral similarity measure named Spectral Correlation Coefficient of Spectrum, i.e., SCC. SCC is the utilization of Pearson's correlation coefficient $r$ in the hyperspectral field and represents the cosine of the generalized angle of tokens after the spectral curves minus their averages, which can be obtained by:

\begin{equation}
\begin{split}
    r(q,k) = \frac{(q-\overline{q})(k-\overline{k})^T}{||(q - \overline{q})|| \cdot ||(k - \overline{k})||},
  \label{eq:SCC}
  \end{split}
\end{equation}

where $\overline{q}, \overline{k}$ denote the mean value of the any two token vectors $q,k$ in $Q,K$. $r \in [-1,1]$ represents whether the correlation degree of $q$ and $k$ are positive or negative. Following the practice in \cite{ding2015novel}, we propose to use $r^2$ to ensure the non-negativity value and regard it as the attention matrix to represent the relationship between any two tokens. Then we have the following theorem for the inductive biases.

\begin{theorem} 
\label{theorem:SCC}
Let $q,k$ be vectors $\in R^{1\times C}$ and $r^2$ be the correlation degree that describes the relationship between any two token vectors, we have the channel-wise translational inductive biases, i.e., scales and shifts, in $r^2$:
\begin{equation}
\begin{split}
  r^2{(q,s \cdot k + t)} &= (\frac{(q-\overline{q})(s \cdot k + t - (s \cdot \overline{k} + t)^T}{||q - \overline{q}|| \cdot ||s \cdot k +t - (s \cdot \overline{k} + t)||})^2
  \\ &= r^2{(q,k)}
  \label{eq:inductive bias}
  \end{split}
\end{equation}
for any $s\in R, t\in R^{1\times C}$.
\end{theorem}

Such characteristics indicate that SCC self-attention is affected by neither shadows nor occlusions that usually lead to scales and offsets transformation in HSIs, where

\begin{equation}
    {\rm SCC}(Q,K,V) = r^2(Q, K)V.
\label{eq:SCC attention}
\end{equation}

Such inductive biases improve the model convergence and make it easy to train on small datasets from scratch. Although the operations on local features, such as pixel unshuffle, limit the receptive field and tend to produce blocking artifacts around the `block' boundaries, as demonstrated in \cite{wang2020deep}, the attention mechanism can effectively enlarge the receptive field by building the long-range dependencies among feature maps, which thus helps to produce natural and smooth images.

\subsection{Efficient SCC-kernel-based self-attention}
Based on SCC self attention, we introduce the proposed efficient SCC-kernel-based self-attention (ESSA) to relieve the computation burden in attention and the calculation process is shown in Figure~\ref{fig:ESSA}. In a kernel machine, in order to get the results of $\psi(Q)\cdot\psi(K)$, it is common to find the kernel function $\mathcal{K}(Q,K)=\psi(Q)\cdot\psi(K)$ so that we can directly calculate $\mathcal{K}(Q,K)$ instead of calculating $\psi(Q)$ and $\psi(K)$ separately, which is known as the kernel trick. In contrast, to lower the computation cost of SCC self-attention, a solution is finding the mapping function $\psi(Q)$ and $\psi(K)$ so that $\psi(Q)\psi(K)=r^2(Q,K)$. Then Equation~\ref{eq:SCC attention} can be reformulated as $\psi(Q)(\psi(K)^TV)$ and decreases from quadratic complexity to linear complexity regarding the token sequence $N$. Before finding the mapping function, we first develop SCC into a radial basis function (RBF)-like kernel \cite{scholkopf2002learning} for non-linearity and good derivatives:

\begin{equation}
  \mathcal{K}_{SCC} = \exp (r^2).
  \label{eq:6}
\end{equation}
In the following, we demonstrate that the SCC-kernel $\mathcal{K}_{SCC}$ is a Mercer's kernel to confirm the existence of the mapping function $\psi()$.
\begin{theorem} 
\label{theorem:Mercer's theorem}
(Mercer's theorem \cite{mercer1909xvi}) Let $\mathcal{X},\mathcal{Y}$ be the input space, and $\mathcal{H}$ be the Hilbert space. If the mapping $\psi(x): \mathcal{X} \longrightarrow \mathcal{H}$ exists, then there is a kernel function $\mathcal{K}(x,y)$ satisfied:
\begin{equation}
  \mathcal{K}(x,y) = \left \langle \psi(x),\psi(y) \right \rangle
  \label{eq:7}
\end{equation}
\end{theorem}

\begin{theorem} 
\label{theorem:Mercer's kernel}
(Mercer's kernel closure properties \cite{burges1998tutorial}) Let $\mathcal{X},\mathcal{Y}$ be the input space and $\mathcal{K}_1$,$\mathcal{K}_2$ be the Mercer's kernels, then:
\begin{itemize}
\item if $\mathcal{K}(x,y) = \mathcal{K}_1(x,y)\mathcal{K}_2(x,y)$, then $\mathcal{K}$ is Mercer's kernel.
\item if $a,b>0$ and $\mathcal{K}(x,y) = a\mathcal{K}_1(x,y) + b\mathcal{K}_2(x,y)$, then $\mathcal{K}$ is Mercer's kernel.
\end{itemize}
\end{theorem}

According to Theorem~\ref{theorem:Mercer's theorem}, we can conclude $r$ is a normalized linear kernel and also a Mercer’s kernel. Mathematically, the Taylor expansion of $\exp(r^2)$ can be expressed as $\exp(r^2) = 1 + r^2 + \frac{(r^2)^2}{2!} + \frac{(r^2)^3}{3!} + \cdots$. According to Theorem~\ref{theorem:Mercer's kernel}, we can easily find $r^2$ and $\exp(r^2)$ Mercer's kernels. It guarantees the existence of the mapping function $\psi()$, which can be acquired via Taylor expansion as follows:

\begin{equation}
\begin{split}
    \mathcal{K}_{SCC}(q,k) & = \exp(\frac{q_{norm}^2 k_{norm}^2}{\sigma} )\\
    & = \sum\limits_{i=0}^\infty\frac{(q_{norm}^2 k_{norm}^2)^i}{\sigma^{i}i!} \\
    & = \sum\limits_{i=0}^\infty(\frac{q_{norm}^{2i}}{\sigma^{\frac{1}{2}i}\sqrt{i!}}\frac{k_{norm}^{2i}}{\sigma^{\frac{1}{2}i}\sqrt{i!}}) \\
    & = \left \langle \psi(q),\psi(k) \right \rangle
\end{split}
\label{eq:8}
\end{equation}

where $q_{norm} = q - \overline{q}$, $k_{norm} = k - \overline{k}$ and $\psi(q) = (1,\frac{q_{norm}^2}{\sigma^\frac{1}{2}},\frac{q_{norm}^4}{2^{\frac{1}{2}} \sigma },\cdots, )$.
We choose the order of the polynomial, i.e., the number of terms, to balance the performance and computation cost during experiments. Finally, the calculation of ESSA is given by

\begin{equation}
\begin{aligned}
  ESSA(Q,K,V) = (\psi(Q) \psi(K)^T) V 
              = \psi(Q) (\psi(K)^T V)
\end{aligned}
\label{eq:ESSA}
\end{equation}

By exchanging the multiplication order, the total computation complexity of ESSA is $\mathcal{O}(NC^2)$, which is significantly smaller than conventional attention $\mathcal{O}(N^2C)$ because the channel dimension is usually much smaller than the sequence length, especially for high-resolution images in HSI-SR. It is noted that the mathematical formulation of ESSA has a significant difference from channel-wise attention \cite{zamir2022restormer}, \ie, $Q\times{\rm softmax}(K^TV)$, demonstrating the different motivations and behaviors between the two methods.

\section{Experiments}

\subsection{Datasets and settings}
\noindent\textbf{Chikusei dataset}: the Chikusei dataset \cite{yokoya2016airborne} is acquired using the Headwall Hyperspec-VNIR-C imaging sensor over agricultural and urban areas in Chikusei, Ibaraki, Japan. The dataset comprises images of 19 different classes, which are collected via a field survey and visual inspection, along with high-resolution images that are captured concurrently with the hyperspectral data.

\noindent\textbf{Cave dataset}: the Cave dataset \cite{yasuma2010generalized} is a multispectral dataset that comprises 32 images of everyday objects. The dataset contains full spectral resolution reflectance data from 400 nm to 700 nm at a resolution of 10 nm, resulting in a total of 31 bands. The images have a resolution of 512 $\times$ 512 pixels and are stored as 16-bit grayscale PNG images per band. 

\noindent\textbf{Pavia dataset}: the Pavia Dataset \cite{gamba2004collection} is a hyperspectral dataset acquired by the ROSIS sensor during a flight campaign over Pavia, northern Italy. The images contain 102 spectral bands with a spatial resolution of 1096 $\times$ 1096, with a geometric resolution of 1.3 m. The image feature area is divided into 9 categories, each containing 9 samples. 

\noindent\textbf{Harvard dataset}: the Harvard dataset \cite{chakrabarti2011statistics} collects fifty indoor and outdoor scenes real-world images. The images are taken from a hyperspectral camera (Nuance FX, CRI Inc.) with wavelengths ranging from 420nm to 730nm at steps of 10nm. Each image has a spatial resolution of 1392 × 1040 with thirty-one spectral measurements at each pixel. 

\noindent\textbf{Implementation details:} We trained our ESSAformer model from scratch using PyTorch and the Adam optimizer. The loss function is L1 loss. We set the initial learning rate to $1\times10^{-4}$ and gradually decreased it to a minimum of $1\times10^{-5}$. We used the same training settings for all four datasets (Chikusei, Pavia, Cave, Harvard), without any special tuning. Our ESSAformer model consists of five stages, each with one encoder layer, and an upsampling module in the last stage. The first $3\times3$ convolution layer projects the channel dimension to $C=256$, which is maintained in all stages except the last $3\times3$ convolution layer that recovers the original channel size. The temperature $\sigma$ is set to 1. We used NVIDIA RTX 3090 GPUs for all experiments.

\paragraph{Evaluation metrics:} We use six popular metrics for all experiments to thoroughly evaluate the model's performance in both spatial and spectral perspectives: the peak signal-to-noise ratio (PSNR), spectral angle mapper (SAM) \cite{yuhas1992discrimination}, erreur relative globale adimensionnelle de synthese (ERGAS) \cite{wald2002data}, structure similarity (SSIM) \cite{wang2004image}, root mean square error (RMSE), and cross correlation (CC) \cite{loncan2015hyperspectral}.

\begin{table}[t]
\setlength{\tabcolsep}{0.8mm}
\centering
\resizebox{\columnwidth}{!}{%
\begin{tabular}{c|c|cccccc|c}
\toprule
\multicolumn{2}{c|}{Method} & MPSNR$\uparrow$ & SAM$\downarrow$ & ERGAS$\downarrow$ & MSSIM$\uparrow$ & RMSE$\downarrow$ & CC$\uparrow$ & MACs(G) \\ 
\midrule
{Bicubic} &  & 43.2125 & 1.7880 & 3.5981 & 0.9721 & 0.0082 & 0.9781 & N/A \\
{GDRRN \cite{li2017hyperspectral}} & & 46.5412 & 1.3779 & 2.5896 & 0.9872 & 0.0055 & 0.9884 & 1.66  \\
{SSPSR \cite{jiang2020learning}} & & 47.4403 & 1.2072 & 2.2805 & 0.9897 & 0.0050 & 0.9910 & 23.47  \\
{MCNet \cite{li2020mixed}} & 2$\times$ & 46.7882 & 1.3311 & 2.4382 & 0.9872 & 0.0055 & 0.9893 & 82.77 \\
{Bi-3DQRNN \cite{fu2021bidirectional}} & & 45.7107 & 1.4306 & 2.7407 & 0.9843 & 0.0061 & 0.9867 & 30.24 \\
{DCM-NET \cite{zhang2021difference}} & & 48.0238 & 1.1160 & 2.1766 & 0.9906 & 0.0047 & 0.9916 & 101.3  \\
{Ours} & & \textbf{48.2886} &\textbf{1.1004} & \textbf{2.1268} & \textbf{0.9912} & \textbf{0.0045} & \textbf{0.9920} & 12.82  \\ 
\midrule
{Bicubic} & & 37.6377 & 3.4040 & 6.7564 & 0.8954 & 0.0156 & 0.9212 & N/A  \\
\multicolumn{1}{l|}{GDRRN \cite{li2017hyperspectral}} & & 39.6456 & 2.6306 & 5.3946 & 0.9353 & 0.0122 & 0.9490 & 6.65 \\
{SSPSR \cite{jiang2020learning}} & & 40.3612 & 2.3527 & 4.9894 & 0.9413 & 0.0114 & 0.9565 & 42.44 \\
{MCNet \cite{li2020mixed}} & 4$\times$ & 39.5599 & 2.7831 & 5.3687 & 0.9317 & 0.0126 & 0.9481 & 289.63 \\
{Bi-3DQRNN \cite{fu2021bidirectional}} & & 39.8938 & 2.5221 & 5.1923 & 0.9377 & 0.0120 & 0.9518 & 120.97 \\
{DCM-NET \cite{zhang2021difference}} & & 40.5139 & 2.3012 & 4.8584 & 0.9464 & 0.0112 & 0.9581 & 130.9 \\
{Ours} & & \textbf{40.7648} & \textbf{2.2126} & \textbf{4.7231} & \textbf{0.9487} & \textbf{0.0109} & \textbf{0.9601} & 48.65 \\ 
\midrule
{Bicubic} & & 34.5049 & 5.0436 & 9.6975 & 0.8069 & 0.0224 & 0.8314 & N/A \\
{GDRRN \cite{li2017hyperspectral}} & & 35.2210 & 4.6363 & 9.0720 & 0.8354 & 0.0202 & 0.7977 & 26.62 \\
{SSPSR \cite{jiang2020learning}} & & 35.8279 & 4.0282 & 8.3177 & 0.8538 & 0.0192 & 0.8773 & 118.33 \\
{MCNet \cite{li2020mixed}} & 8$\times$ & 35.2643 & 4.6107 & 8.7438 & 0.8321 & 0.0208 & 0.8588 & 2637.51 \\
{Bi-3DQRNN \cite{fu2021bidirectional}} & & 35.6284 & 4.2259 & 8.4955 & 0.8456 & 0.0196 & 0.8701 & 483.89 \\
{DCM-NET \cite{zhang2021difference}} & & 35.9809 & 3.9310 & 8.1459 & 0.8580 & 0.0189 & 0.8811 & 249.95 \\
{Ours} & & \textbf{36.1405} & \textbf{3.8979} & \textbf{8.1181} & \textbf{0.8599} & \textbf{0.0187} & \textbf{0.8823} & 192.64 \\ 
\bottomrule 
\end{tabular}
}
\caption{Quantitative comparison of different methods on the Chikusei dataset.}
\label{tab:chikusei}
\end{table}

\begin{figure}[]
\centering
\includegraphics[width=\linewidth]{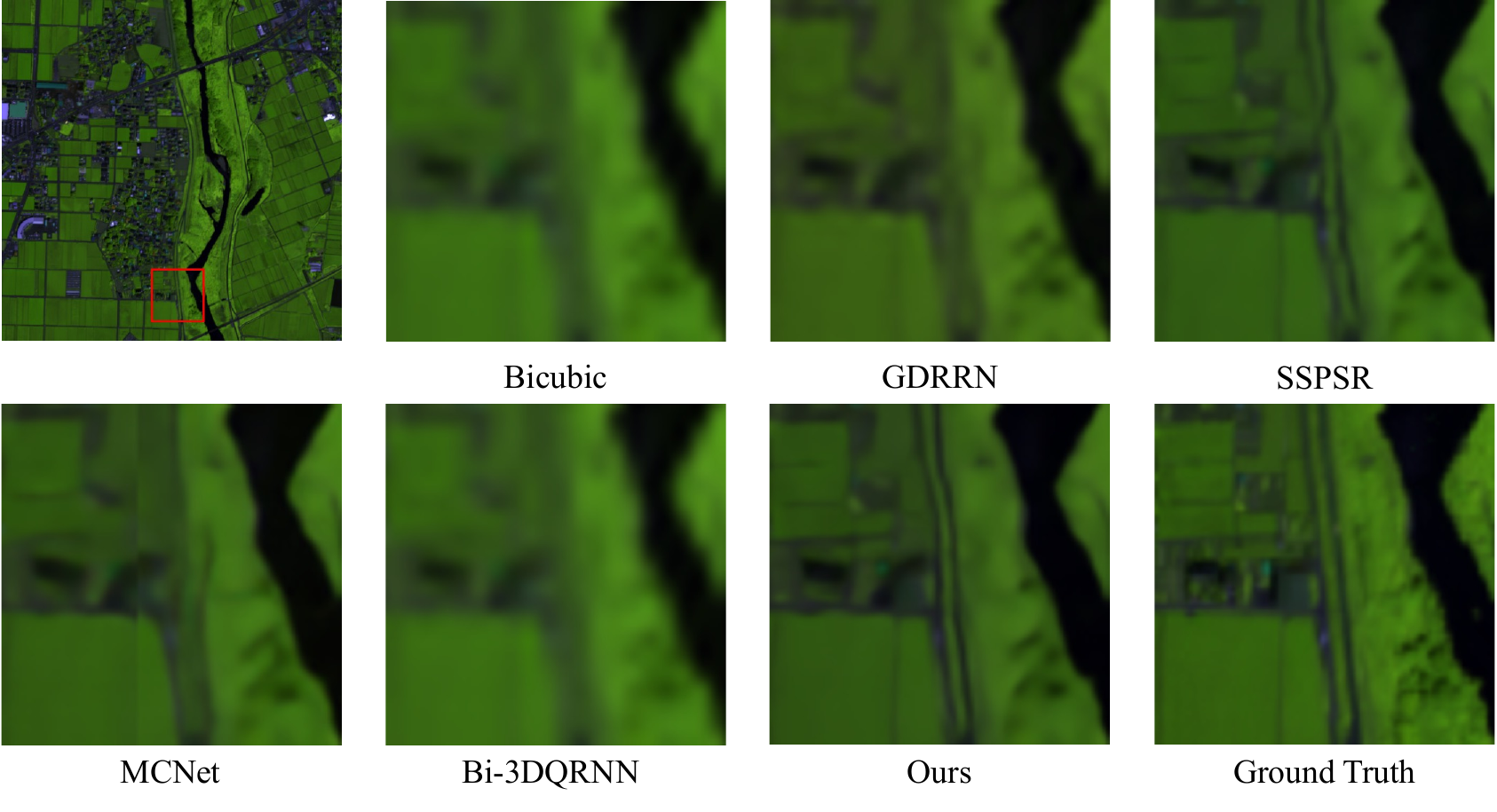}
\caption{Chikusei's visual results, i.e., the patch in red rectangle, of different methods are provided for comparison. We set the bands of 70/100/36 as the R/G/B channels for better visualization.  }

\label{fig:chikusei}
\end{figure}

\subsection{Qualitative results}
We compare our methods with five representative deep learning methods includes GDRRN \cite{li2017hyperspectral}, SSPSR \cite{jiang2020learning}, MCNet \cite{li2020mixed}, Bi-3DQRNN \cite{fu2021bidirectional}, and DCM-NET \cite{zhang2021difference}, besides traditional bicubic interpolation. All the models are trained from scratch. The qualitative and visual results will be presented below by datasets:
\paragraph{Experiments on the Chikusei dataset:} 
Images in Chikusei have 2517 $\times$ 2335 pixels with 128 bands.
We follow SSPSR \cite{jiang2020learning} to crop non-overlapped patches of 512 $\times$ 512 resolution. For each image in Chikusei, four cropped patches are used for testing and the rest are for training. We have three scale factors for experiments. Specifically, the input resolutions for scale factors of 2, 4, and 8 are set to $32\times32$, $16\times16$, and $16\times16$, respectively. The output resolutions are $64\times64$, $64\times64$, and $128\times128$, respectively.

The quantitative results for the different methods are demonstrated in Table~\ref{tab:chikusei} with the best performance highlighted in bold. Our ESSA outperforms other approaches in almost all metrics at all three scale factors, demonstrating the effectiveness of our model. For example, ESSAformer outperforms the second, i.e., DCM-Net \cite{zhang2021difference}, by 0.26 dB in PSNR and 0.13 in ERGAS for the $4\times$ scale factor. To further measure the performance of our model, the visual results are presented in Figure~\ref{fig:chikusei}. We zoomed in the area in red rectangle and provide the details produced by each method. It can be seen that only SSPSR, DCM-NET, and our method recovered the two vertical lines of the original image clearly, and among these three methods, SSPSR yields the worst result whose reconstructed lines are broken in many places. DCM-NET recovers relatively better than SSPSR, but discontinuation also appears, and our ESSA produces the best results.

\begin{table*}[htbp]
  \centering
  \resizebox{\textwidth}{!}{%
    \begin{tabular}{c|c|cccccc|cccccc}
    \toprule
    \multicolumn{2}{c|}{\multirow{2}[3]{*}{Method}} & \multicolumn{6}{c|}{Pavia}                     & \multicolumn{6}{c}{Cave} \\
\cmidrule{3-14}    \multicolumn{2}{c|}{} & MPSNR$\uparrow$ & SAM$\downarrow$ & ERGAS$\downarrow$ & MSSIM$\uparrow$ & RMSE$\downarrow$ & CC$\uparrow$ & MPSNR$\uparrow$ & \multicolumn{1}{c}{SAM$\downarrow$} & ERGAS$\downarrow$ & MSSIM$\uparrow$ & RMSE$\downarrow$ & CC$\uparrow$ \\
    \midrule
    Bicubic &       & 34.4107 & 5.2881 & 4.4877 & 0.9387 & 0.0197 & 0.9760 & \multicolumn{1}{c}{38.0603} & 3.237 & \multicolumn{1}{c}{4.9579} & \multicolumn{1}{c}{0.9662} & \multicolumn{1}{c}{0.0147} & \multicolumn{1}{c}{0.9907} \\
    GDRRN \cite{li2017hyperspectral} &       & 37.4868 & 4.7773 & 3.2812 & 0.9651 & 0.0138 & 0.9867 & \multicolumn{1}{c}{40.9785} & 3.7454 & \multicolumn{1}{c}{4.2106} & \multicolumn{1}{c}{0.9738} & \multicolumn{1}{c}{0.0126} & \multicolumn{1}{c}{0.9948} \\
    SSPSR \cite{jiang2020learning} &       & 37.7264 & 4.6532 & 3.1757 & 0.9668 & 0.0135 & 0.9875 & \multicolumn{1}{c}{41.3895} & 3.1472 & \multicolumn{1}{c}{3.3333} & \multicolumn{1}{c}{0.9752} & \multicolumn{1}{c}{0.0101} & \multicolumn{1}{c}{0.9953} \\
    MCNet \cite{li2020mixed} & \multicolumn{1}{c|}{2$\times$} & 37.2858 & 4.7874 & 3.3173 & 0.9638 & 0.0143 & 0.9864 & \multicolumn{1}{c}{41.9772} & 2.6656 & \multicolumn{1}{c}{3.1483} & \multicolumn{1}{c}{0.9765} & \multicolumn{1}{c}{0.0095} & \multicolumn{1}{c}{0.9956} \\
    Bi-3DQRNN \cite{fu2021bidirectional} &       & 36.9049 & 4.7343 & 3.4460 & 0.9623 & 0.0148 & 0.9855 & \multicolumn{1}{c}{40.7212} & 2.8859 & \multicolumn{1}{c}{3.6087} & \multicolumn{1}{c}{0.9744} & \multicolumn{1}{c}{0.0108} & \multicolumn{1}{c}{0.9945} \\
    DCM-NET \cite{zhang2021difference} &       & 37.1815 & 5.2427 & 3.3738 & 0.9600 & 0.0144 & 0.9861 & \multicolumn{1}{c}{41.9867} & 2.7051 & \multicolumn{1}{c}{3.1217} & \multicolumn{1}{c}{0.9771} & \multicolumn{1}{c}{0.0095} & \multicolumn{1}{c}{0.9957} \\
    Ours  &       & \textbf{38.7896} & \textbf{4.5576} & \textbf{2.8806} & \textbf{0.9712} & \textbf{0.0120} & \textbf{0.9896} & \textbf{42.2174} & \multicolumn{1}{c}{\textbf{2.6623}} & \textbf{3.0443} & \textbf{0.9778} & \textbf{0.0092} & \textbf{0.9958} \\
    \hline
    Bicubic &        & 29.6732 & 6.9353 & 7.5858 & 0.8154 & 0.0347 & 0.9321 & \multicolumn{1}{c}{33.0421} & 4.7962 & \multicolumn{1}{c}{7.846} & \multicolumn{1}{c}{0.9202} & \multicolumn{1}{c}{0.0258} & \multicolumn{1}{c}{0.9767} \\
    GDRRN \cite{li2017hyperspectral} &        & 30.8474 & 6.5915 & 6.7641 & 0.8677 & 0.0302 & 0.9477 & \multicolumn{1}{c}{34.897} & 4.3822 & \multicolumn{1}{c}{6.7552} & \multicolumn{1}{c}{0.938} & \multicolumn{1}{c}{0.0206} & \multicolumn{1}{c}{0.9830}\\
    SSPSR \cite{jiang2020learning} &        & 30.6447 & 6.4081 & 6.776 & 0.8619 & 0.0312 & 0.9461 & \multicolumn{1}{c}{35.3433} & 4.1654 & \multicolumn{1}{c}{6.5045} & \multicolumn{1}{c}{0.9434} & \multicolumn{1}{c}{0.0200}  & \multicolumn{1}{c}{0.9838}\\
    MCNet \cite{li2020mixed} & \multicolumn{1}{c|}{4$\times$}  & 30.9330 & 6.6822 & 6.5725 & 0.8628 & 0.0299 & 0.9488 & \multicolumn{1}{c}{35.5813} & \textbf{3.7189} & \multicolumn{1}{c}{6.3928} & \multicolumn{1}{c}{\textbf{0.9470}} & \multicolumn{1}{c}{0.0194} & \multicolumn{1}{c}{0.9845}\\
    Bi-3DQRNN \cite{fu2021bidirectional} &        & 30.4242 & 7.2323 & 7.0308 & 0.8525 & 0.0317 & 0.9417 & \multicolumn{1}{c}{35.3269} & 3.9294 & \multicolumn{1}{c}{6.5136} & \multicolumn{1}{c}{0.9438} & \multicolumn{1}{c}{0.0199} & \multicolumn{1}{c}{0.9839}\\
    DCM-NET \cite{zhang2021difference} &        & 30.6048 & 7.2623 & 6.8190 & 0.8507 & 0.0312 & .9454 & \multicolumn{1}{c}{35.5055} & 3.9460 & \multicolumn{1}{c}{6.4092} & \multicolumn{1}{c}{0.9445} & \multicolumn{1}{c}{0.0197} & \multicolumn{1}{c}{0.9957}\\
    Ours  &        & \textbf{31.6126} & \textbf{6.3032} & \textbf{6.1063} & \textbf{0.8847} & \textbf{0.0275} & \textbf{0.9558} & \textbf{35.8947} & \multicolumn{1}{c}{3.8390} & \textbf{6.1621} & 0.9467 & \textbf{0.0190} & \textbf{0.9869} \\
    \hline
    Bicubic &        & 26.6043 & 8.4814 & 10.7741 & 0.6509 & 0.0500 & 0.8597 & \multicolumn{1}{c}{29.2466} & 6.6079 & \multicolumn{1}{c}{12.2687} & \multicolumn{1}{c}{0.832} & \multicolumn{1}{c}{0.039} & \multicolumn{1}{c}{0.9439}\\
    GDRRN \cite{li2017hyperspectral} &        & 26.7832 & 9.2154 & 10.5724 & 0.6718 & 0.0488 & 0.8698 & \multicolumn{1}{c}{30.3026} & 7.151 & \multicolumn{1}{c}{11.0352} & \multicolumn{1}{c}{0.8513} & \multicolumn{1}{c}{0.0353} & \multicolumn{1}{c}{0.9533}\\
    SSPSR \cite{jiang2020learning} &        & 26.8435 & 10.1753 & 10.4754 & 0.6692 & 0.0487 & 0.9595 & \multicolumn{1}{c}{31.129} & 5.5101 & \multicolumn{1}{c}{10.1804} & \multicolumn{1}{c}{0.8749} & \multicolumn{1}{c}{0.0325} & \multicolumn{1}{c}{0.9595}\\
    MCNet \cite{li2020mixed} & \multicolumn{1}{c|}{8$\times$}  & 27.0388 & 8.7111 & 10.2412 & 0.6808 & 0.0475 & 0.8734 & \multicolumn{1}{c}{31.323} & 5.7398 & \multicolumn{1}{c}{10.0206} & \multicolumn{1}{c}{0.8804} & \multicolumn{1}{c}{0.0320} & \multicolumn{1}{c}{0.9607}\\
    Bi-3DQRNN \cite{fu2021bidirectional} &        & 26.9813 & \textbf{8.4730} & 10.312 & 0.6802 & 0.0479 & 0.8715 & \multicolumn{1}{c}{31.1791} & 5.3401 & \multicolumn{1}{c}{10.1370} & \multicolumn{1}{c}{0.8792} & \multicolumn{1}{c}{0.0322} & \multicolumn{1}{c}{0.9601}\\
    DCM-NET \cite{zhang2021difference} &        & 25.6571 & 13.4698 & 12.014 & 0.5487 & 0.0559 & 0.8331 & \multicolumn{1}{c}{31.3766} & 5.3067 & \multicolumn{1}{c}{9.9363} & \multicolumn{1}{c}{0.8822} & \multicolumn{1}{c}{0.0316} & \multicolumn{1}{c}{0.9618}\\
    Ours  &        & \textbf{27.1114} & 8.9197 & \textbf{10.1681} & \textbf{0.6918} & \textbf{0.0471} & \textbf{0.8757} & \textbf{31.4387} & \textbf{5.3041} & \textbf{9.9058} & \textbf{0.8845} & \textbf{0.0316} & \textbf{0.9629}\\
    \bottomrule
    \end{tabular}%
  }
  \caption{Quantitative comparison of different methods on the Cave and
Pavia datasets. The results of different scales are given respectively, where the best performance of each scale is highlighted in bold.}
  \label{tab:paviaandcave}%
\end{table*}%

\begin{figure}[]
\centering
\includegraphics[width=\linewidth]{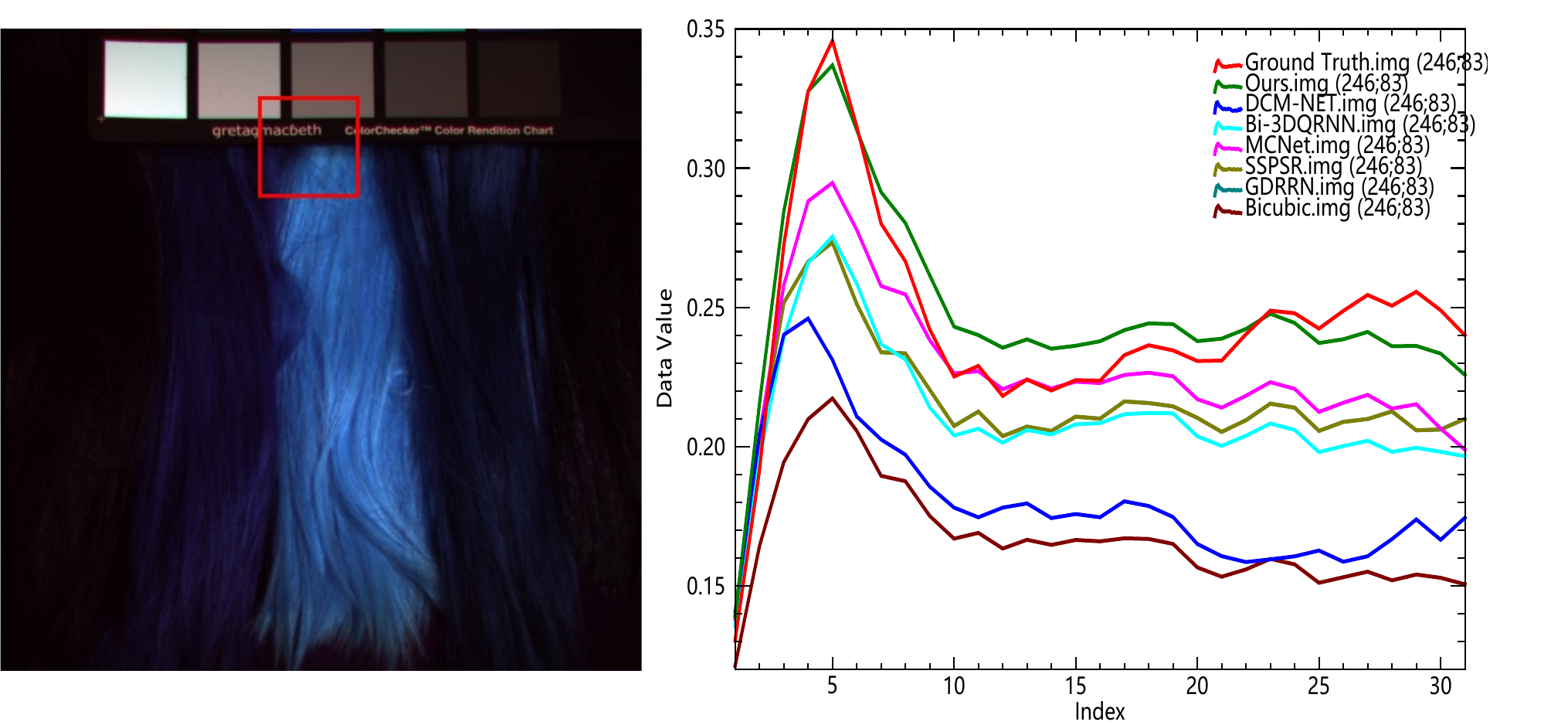}
\caption{The test image in Cave and the spectral profile of the area in red rectangle are provided respectively. The red line refers to the ground truth and the closest green line refers to the ESSAformer's result, which shows the strong ability of the ESSAformer to recover the texture detail.}
\label{fig:cave}
\end{figure}

\paragraph{Experiments on the Cave dataset:} The Cave dataset \cite{yasuma2010generalized} has 32 images with 512 $\times$ 512 resolution of different scenes with full spectral resolution reflectance data from 400 to 700 nm at a resolution of 10 nm (31 bands in total). We randomly chose 8 scenes for testing and the remained images are for training. The same cropping settings of Chikusei are used for experiments with Cave.

The results are represented in Table~\ref{tab:paviaandcave}. As can be seen, our network still obtains the best performance in most metrics and scale factors, which confirms ESSAformer's superiority. Besides, we compare the spectral profile of the images from different methods in Figure~\ref{fig:cave}. As can be seen in right right image of Figure~\ref{fig:cave}, our method, i.e., the green line, is closest to the groundtruth (the red line) and recovers better details, while the other methods cannot restore the peak value as well as ours.

\begin{figure}[]
\centering
\includegraphics[width=\linewidth]{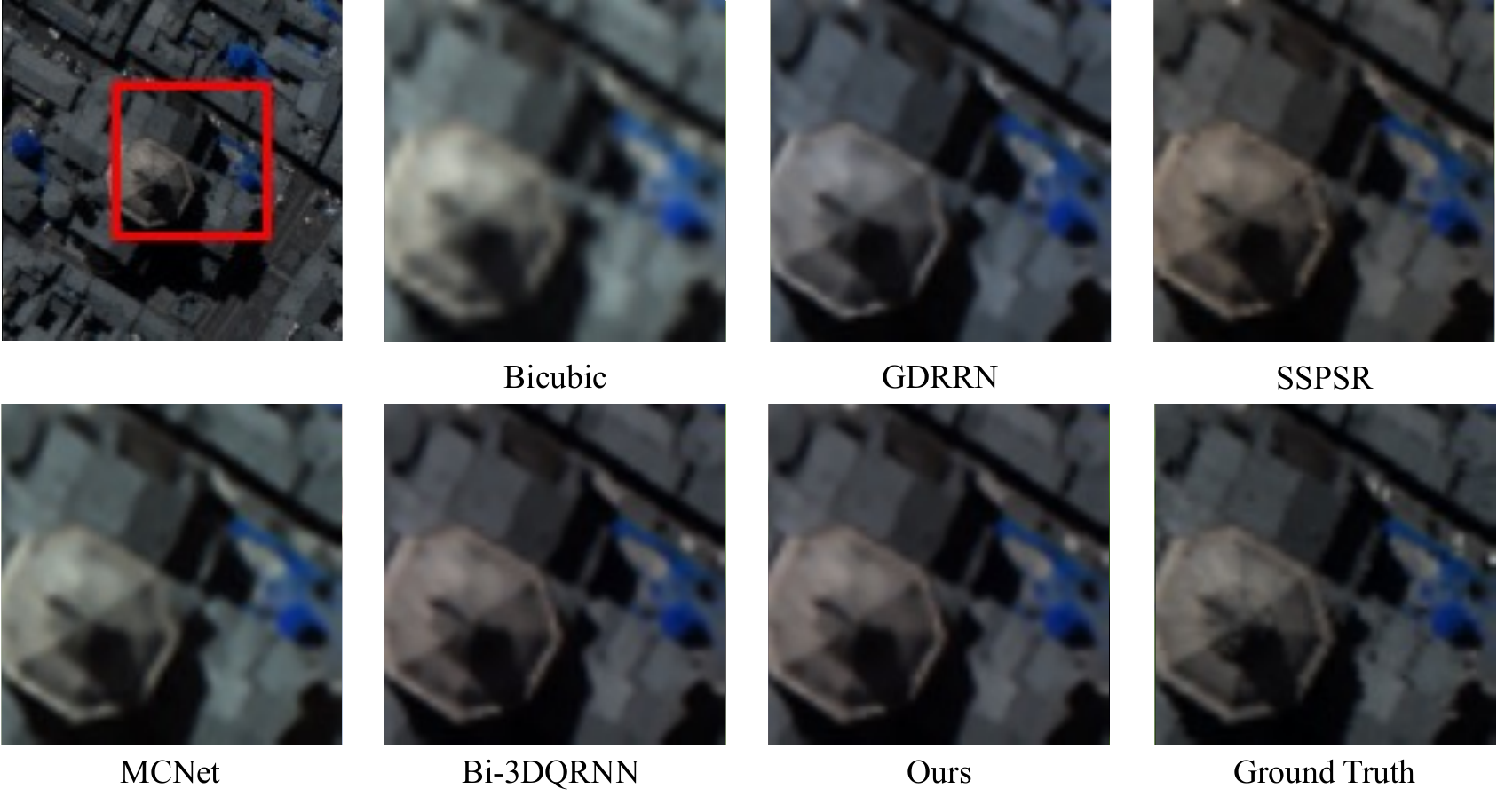}
\caption{The test image from Pavia dataset and details of the area in red rectangle offered by various methods. Bands 10/70/50 are visualized as the R/G/B channels.}
\label{fig:pavia}
\end{figure}

\paragraph{Experiments on the Pavia dataset:} The size of Pavia is significantly small compared with Chikusei and Cave. It has only one $1096\times 1096$ image with the available part $1096\times 714$. Similarly, we crop non-overlapped patches with a spatial resolution of $120\times 714$. For each image in Pavia, three patches are used for testing and the rest for training.

As can be seen in Table~\ref{tab:paviaandcave}, ESSAformer reaches the best performance and significantly outperforms other methods in most metrics. For example, it obtains 38.79 dB and 31.61 dB in PSNR and has more than 1 dB gain when comparing the second. ESSAformer obtains the best 0.0275 and 0.955 for RMSE and CC, respective. This confirms the effectiveness of the proposed techniques for small datasets, i.e., the channel-wise inductive bias in ESSA.  It is noteworthy that the performance decreases when the scale factor rises from 2$\times$ to $8\times$ because of the increased task difficulty. 

\begin{table}[htbp]
  \centering
  \caption{Comparison of different methods on Harvard dataset.}
  \footnotesize
    \begin{tabular}{c|cc|cc}
    \hline
    \multirow{2}[1]{*}{Methods} & \multicolumn{2}{c|}{Harvard 2$\times$} & \multicolumn{2}{c}{Harvard 4$\times$} \\
\cline{2-5}    & PSNR $\uparrow$ & SAM $\downarrow$ & PSNR $\uparrow$ & SAM $\downarrow$ \\
    \hline
    DCM-NET \cite{zhang2021difference} & 50.2559 & 2.7389 & 45.4087 & 3.3102 \\
    SSPSR \cite{jiang2020learning} & 50.2929 & 2.7017 & 45.2164 & 3.4292 \\
    \textbf{Ours} & \textbf{50.6928} & \textbf{2.6489} & \textbf{45.5091} & \textbf{3.3031} \\
    \hline
    \end{tabular}%
  \label{tab:harvard}%
\end{table}%

\paragraph{Experiments on the Harvard dataset:} In addition to the aforementioned three datasets, we conduct experiments on the Harvard dataset \cite{chakrabarti2011statistics} and compare ESSAFormer with other methods. We follow the training settings in \cite{zhang2021difference} for our method and present the results in Table~\ref{tab:harvard}. It can be seen that ESSAFormer outperforms the others by a significant margin regarding both PSNR and SAM metrics, demonstrating the effectiveness of our proposed model.

\begin{table}[htpb]
\centering
\resizebox{\columnwidth}{!}{%
\begin{tabular}{@{}cccccccccc@{}}
\toprule
Method    & CNN     & MHSA\cite{vaswani2017attention} & Swin\cite{liu2021swin} & ESSA(0) & \multicolumn{2}{c}{ESSA(1)} & ESSA(2) & \multicolumn{2}{c}{ESSA(3)} \\ \midrule
PSNR      & 40.5672 & 40.4310                                          & 40.1122                                 & 40.5474 & \multicolumn{2}{c}{40.7648} & 40.7891 & \multicolumn{2}{c}{40.8012} \\
SAM       & 2.2455  & 2.3835                                           & 2.5107                                  & 2.3000  & \multicolumn{2}{c}{2.2126}  & 2.2100  & \multicolumn{2}{c}{2.2106}  \\
SSIM      & 0.9473  & 0.9456                                           & 0.9426                                  & 0.9466  & \multicolumn{2}{c}{0.9487}  & 0.9491  & \multicolumn{2}{c}{0.9495}  \\
MACs(G)   & 62.23   & 77.14                                            & 50.51                                   & 47.12   & \multicolumn{2}{c}{48.65}   & 50.47   & \multicolumn{2}{c}{52.75}   \\
Params(M) & 13.47   & 11.64                                            & 11.64                                   & 11.1    & \multicolumn{2}{c}{11.1}    & 11.1    & \multicolumn{2}{c}{11.1}    \\ \bottomrule
\end{tabular}
}
\caption{The ablation study results of ESSAformer with different attention mechanisms.}
\label{tab:ESSA.}
\end{table}

\begin{table}[t]
\centering
\scriptsize
\begin{tabular}{@{}cclccc@{}}
    \toprule
    Blocks & \multicolumn{2}{c}{PSNR}    & SAM    & SSIM   & MACs(G)  \\ \midrule
    3      & \multicolumn{2}{c}{40.4200} & 2.3703 & 0.9451 & 31.58  \\
    5      & \multicolumn{2}{c}{40.7648} & 2.2126 & 0.9487 & 48.39  \\
    7      & \multicolumn{2}{c}{40.7314} &   2.2359 &  0.9488  &  63.39  \\
    9      & \multicolumn{2}{c}{40.7191} &  2.2279  &  0.9489  &   78.30   \\ \bottomrule
\end{tabular}
%
\caption{The ablation study results regarding the impact of the block number.}
\label{tab:Number of Blocks.}
\end{table}

\begin{table}[t]
\centering
\scriptsize
\begin{tabular}{@{}l|clccc@{}}
\toprule
 & \multicolumn{5}{c}{image resolution}                                                  \\
                     & \multicolumn{2}{c}{10 $\times$ 10} & 20 $\times$ 20 & 30 $\times$ 30 & 40 $\times$ 40 \\ 
\hline
MHSA                 & \multicolumn{2}{c}{23.67 G}        & 142.09 G       & 494.70 G       & 1326.80 G      \\
ESSA                 & \multicolumn{2}{c}{19.06 G}        & 76.22 G        & 173.66 G       & 304.97 G       \\ \bottomrule
\end{tabular}
\caption{Computation cost (FLOPs) comparison between MHSA and ESSA. The scale factor is $4\times$ and the channel dimension is 128.}
\label{tab:Model Efficiency.}
\end{table}

\begin{table}[htbp]
\centering
\resizebox{\columnwidth}{!}{%
\begin{tabular}{ccccccc}
\toprule
          & {GDRRN} & {SSPSR} & {MCNet} & {Bi-3DQRNN} & {DCM-NET} & {Ours}  \\ \midrule
          MPSNR (dB) & 39.65 & 40.36 & 39.56 & 39.89 & 40.51 & 40.76 \\
MACs(G)   & 6.65                       & 42.44                      & 289.63                     & 120.97                         & 130.9                        & {48.65} \\ 
Params(M) & 0.442                      & 14.88                      & 2.17                       & 1.29                           & 12.61                        & {11.1}  \\ 
    Time (per epoch) & 2m23s & 6m00s & 46m52s  & 53m26s & 53m04s & 4m26s\\ \bottomrule
\end{tabular}}
\caption{Model efficiency of different approaches on Chikusei 4$\times$.}
\label{tab:model e}
\end{table}

\begin{table}[htbp]
  \centering
  \resizebox{\linewidth}{!}{
    \begin{tabular}{c|ccccc}
    \hline
    Methods & DCM-NET \cite{zhang2021difference} & MCNet \cite{li2020mixed} & Bi-3DQRNN \cite{fu2021bidirectional} & SSPSR \cite{jiang2020learning} & Performer \cite{yuan2021tokens} \\
    \hline
    FPS (n/s) & 5.58 & 20.00 & 14.12 & 36.75 & 55.74 \\
    \hline
    Methods & NPRF \cite{luo2021stable} & ELAN \cite{zhang2022efficient} & RLFN \cite{kong2022residual} & SwinIR \cite{liang2021swinir} & \textbf{Ours} \\
    \hline
    FPS (n/s) & 71.43    & 51.47 & \textbf{336.70}  & 63.29  & \underline{151.06} \\
    \hline
    \end{tabular}}%
  \caption{Inference speed of various methods on the Chikusei dataset 4$\times$ with an NVIDIA 3090 GPU.}
  \label{tab:time}%
\end{table}%

\subsection{Ablation study}
Several ablation studies are conducted to thoroughly understand the proposed network. All the experiments are based on the Chikusei dataset with a scale factor of 4 unless specifically indicated.

\paragraph{Ablation study on polynomial order in ESSA:}
To verify the effectiveness of ESSA, we compare several attention types with ESSA, and the results are demonstrated in Table~\ref{tab:ESSA.}. `CNN' refers to using two 3 $\times$ 3 convolution layers with interval LeakyRelu to replace ESSA. For `MHSA', we use the original Multi-head Self-Attention. For `Swin', we substitute ESSA with the shifted window self-attention from Swin Transformer. `ESSA(0)', `ESSA(1)', `ESSA(2)', and `ESSA(3)' all denote using ESSA, while the difference is the order of Taylor expansion. `3' means that the term below or equal to the third order of the Taylor expansion is kept to approach the function. For `ESSA(0)', the constant value of 1 is used for the mapping function $\psi()$ and the attention matrix becomes constant.

As can be seen in Table~\ref{tab:ESSA.}, using two approaching terms for the mapping function $\psi()$, i.e., ESSA(1), is significantly better than ESSA(0) and the performance increases from 40.5474dB/2.3/0.9466 to 40.7340dB/2.2283/0.9487 on MPSNR/SAM/MSSIM. When increasing the order of kept terms, the PSNR performance improves at a cost of increased computation. Thus we choose to expand the Taylor polynomial to an order of 1 to balance between performance and computation cost. Thanks to the inductive biases in ESSA, ESSAformer fits HSIs well and has better data efficiency. Therefore, it obtains significantly better performance and less computation than CNN, MHSA, and Swin, which demonstrates the effectiveness of the proposed ESSA method in the super-resolution task.

\paragraph{Ablation study on the number of stages:}
ESSAformer uses shared parameters to save the model size and thus each stage can be regarded as a refinement process. We ablate how the number of stages affects the performance as shown in Table~\ref{tab:Number of Blocks.}, where the results of PSNR, SAM, SSIM, and MACs are given for comparison. When the stage increases from 3 to 5, the performance significantly improves from 40.42 to 40.76 dB in PSNR and the computation cost also rises by 28 G. When the number of stages further increases to 7 and 9, the performance saturates and thus we set 5 stages for the proposed ESSAformer.

\paragraph{Efficiency analysis:}
Due to the $\mathcal{O}(N^2)$ complexity, MHSA handles images with large resolutions in an expensive manner. 
In Table \ref{tab:Model Efficiency.}, we compare the overall computational cost with different image resolutions regarding different attention, i.e., conventional MHSA and our proposed ESSA. We adopt the ESSAformer architecture for fair comparison. It can be seen that the computational cost of ESSA and MHSA is relatively similar for infrequent small images with the resolution 10 $\times$ 10. With the resolution going up, the computational cost of MHSA rises sharply and the difference between MHSA and ESSA increases rapidly. The cost of MHSA is 4 times over ESSA when the input image has $40\times40$ resolution, \ie, 1326 GFlops, a huge computation burden for the computing resources.

To further highlight the efficiency of our ESSA, we have conducted comprehensive experiments comparing the efficiency of various HSI-SR models in Table~\ref{tab:model e}. We also evaluate the inference speed of HSI-SR SOTAs, RGB-SR SOTAs, and linear attention variants in Table~\ref{tab:time}. The results clearly demonstrate that our model excels in striking a balance between computational cost, training speed, and performance when compared to other state-of-the-art methods. This further underscores the effectiveness of ESSAformer.

\begin{figure}[t]
\begin{center}
   \includegraphics[width=0.9\linewidth]{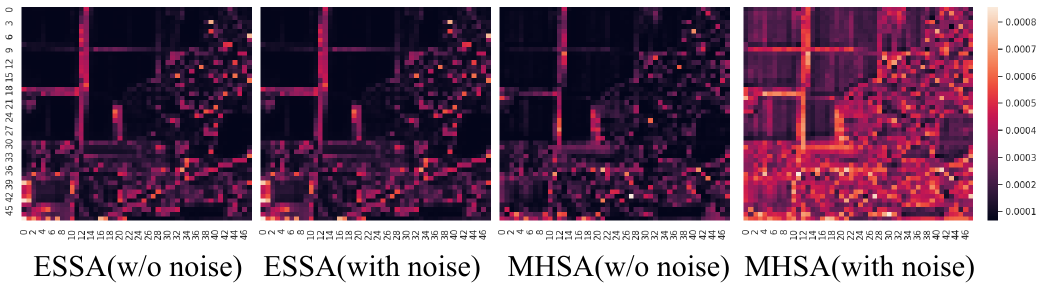}
\end{center}
   \caption{Visualization of the similarity map of ESSA and MHSA before and after noise for comparison.}
\label{fig:noise}
\end{figure}

\begin{figure}[htbp]
    \small
    \begin{minipage}[t]{0.5\linewidth}
        \centering
        \includegraphics[width=0.9\textwidth]{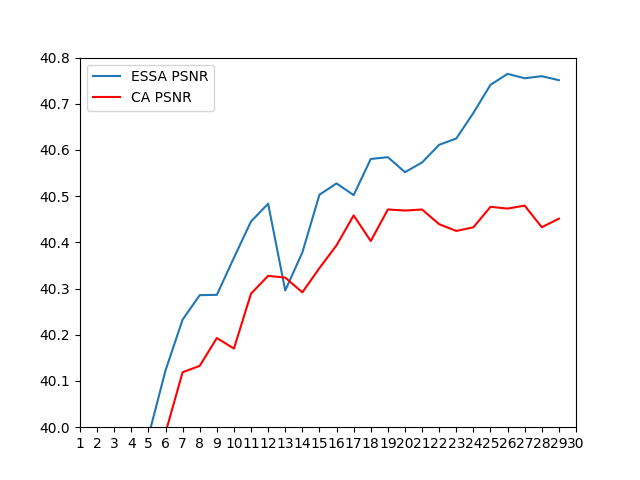}
        \centerline{(a) Performance}
    \end{minipage}%
    \begin{minipage}[t]{0.5\linewidth}
        \centering
        \includegraphics[width=0.9\textwidth]{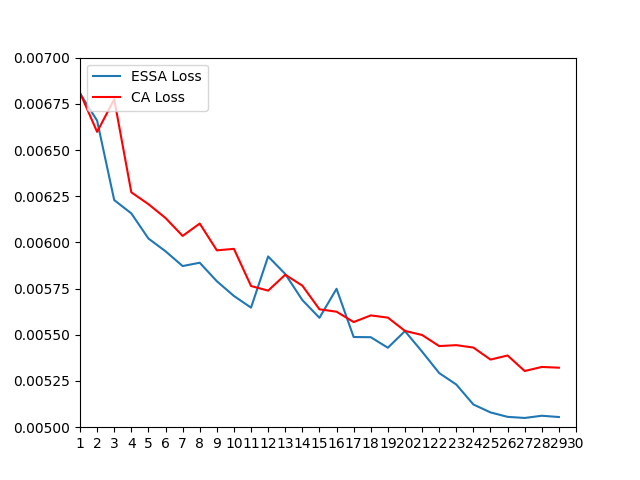}
        \centerline{(b) Train Loss}
    \end{minipage}
    \caption{Comparison between the proposed ESSA and channel-wise attention.}
    \label{fig:ca}
\end{figure}

\noindent\textbf{Comparison to other linear attention.}
In contrast to traditional linear attention targeting RGB images, our ESSA considers the characteristics of the hyperspectral field and excels at handling HSIs. We compare ESSA and other linear attention mechanisms using the Chikusei and Pavia datasets (4$\times$ and show the experimental results in Tab.~\ref{tab:linear}, where all methods use the same model architecture of ESSAFormer except attention for fair comparisons. From the table, we can see that the best performance of ESSA showcases its superiority among all attention choices, underscoring the method's effectiveness in processing HSIs.

\paragraph{Comparisons on large size datasets}1
We perform comparisons on the popular large ICVL \cite{arad_and_ben_shahar_2016_ECCV} and NTIRE \cite{arad2020ntire,arad2022ntire} datasets, whose data volumes are larger than the previous Chikusei and Pavia. According to the results in Tab.~\ref{tab:larger}, although the initial design targets at improving data efficiency from small-scale datasets, ESSAFormer obtains the best performance on ICVL and NTIRE2022, demonstrating its efficacy in learning from large-size datasets. It is noted that ESSAFormer achieves comparable performance to SwinIR on NTIRE2020 while exhibiting more than 2$\times$ faster inference speed as shown in Tab.~\ref{tab:time}, showing a better trade-off between performance and efficiency.

\paragraph{Attention map analysis.}
We compare the similarity maps from the original attention and ESSA in Fig.~\ref{fig:noise}. We consider the first-order results for ESSA due to the infinite Fourier series decomposition. We simulate the occlusions or shadows by using scaling factors and shifts as noise and visualize the resulting heatmaps before and after the simulation. From the figure, we can see ESSA recognizes a highly similar pattern to MHSA. Also, ESSA demonstrates insensitivity to the introduced noise and still focuses on the discriminative features, while MHSA collapses under the same conditions, highlighting the suitability of ESSA for denoising HSIs in the SR task. This merit originates from ESSA's superior designs and accompanies the mathematical principles, \ie, channel-wise translational inductive biases, as proved in \textit{Theorem} \ref{theorem:SCC}. 

\paragraph{Comparison between proposed ESSA and channel-wise attention}
We plot the performance and training loss of ESSA and channel-wise attention \cite{zamir2022restormer,cai2022mask} in Figure~\ref{fig:ca}. The same architecture is adopted for two attentions for a fair comparison. It can be seen that the ESSA has much better training efficiency than the counterpart channel-wise attention, demonstrating the effectiveness of our proposed ESSA mechanism.

\begin{table}[htbp]
  \centering
    \footnotesize
    \begin{tabular}{c|cc|cc}
    \hline
    \multirow{2}[1]{*}{Methods} & \multicolumn{2}{c|}{Chikusei 4$\times$} & \multicolumn{2}{c}{Pavia 4$\times$} \\
\cline{2-5}          & PSNR $\uparrow$  & SAM $\downarrow$   & PSNR $\uparrow$  & SAM $\downarrow$ \\
    \hline
    NPRF \cite{luo2021stable} &    40.4344   &   2.3435    & 31.3306 & 7.372 \\
    Performer \cite{yuan2021tokens} & 40.5833 & 2.2373 & 31.4690 & 6.2658 \\
    Linear \cite{katharopoulos2020transformers} & 40.3824 & 2.4021 &   31.3164    & 6.7403 \\
    \textbf{Ours} & \textbf{40.7648} & \textbf{2.2126} & \textbf{31.6126} & \textbf{6.1063} \\
    \hline
    \end{tabular}%
  \caption{Comparison with different linear attention methods.}
  \label{tab:linear}%
\end{table}%

\begin{table}[htbp]
  \centering
  \footnotesize
    \begin{tabular}{c|c|c|c|c}
    \hline
     & ELAN \cite{zhang2022efficient} & RLFN \cite{kong2022residual} & SwinIR \cite{liang2021swinir} & \textbf{Ours} \\
    \hline
    PSNR $\uparrow$ & 39.5227 & 39.6813 & 39.5165 & \textbf{40.7648} \\
    SAM $\downarrow$ & 2.6945 & 2.6583 & 2.6212 & \textbf{2.2126} \\
    \hline
    \end{tabular}%
    \caption{Comparisons with RGB-SOTA methods on Chikusei 4$\times$.}
  \label{tab:rgb}%
\end{table}%

\begin{table}[htbp]
  \centering
  \resizebox{\linewidth}{!}{
      \scriptsize
    \begin{tabular}{c|cc|cc|cc}
    \hline
    \multirow{2}[1]{*}{Methods} & \multicolumn{2}{c|}{ICVL \cite{arad_and_ben_shahar_2016_ECCV}} & \multicolumn{2}{c|}{NTIRE2020 \cite{arad2020ntire}} & \multicolumn{2}{c}{NTIRE2022 \cite{arad2022ntire}} \\
\cline{2-7}   & PSNR $\uparrow$ & SAM $\downarrow$ & PSNR $\uparrow$ & SAM $\downarrow$ & PSNR $\uparrow$ & SAM $\downarrow$ \\
    \hline
    RLFN \cite{kong2022residual}  & 38.2526 & 2.2179 & 34.2946 & 1.9554 & 37.0744 & 1.3873\\
    SwinIR \cite{liang2021swinir} & 39.0727 & 1.9303 & \textbf{34.9186} & 1.8421 & 37.7432 & 1.4881\\
    DCMNet \cite{zhang2021difference} & 38.7663 & 2.3337 & 34.5316 & 2.0283 & 37.4761 & 1.6257\\
    \textbf{Ours} &   \textbf{39.5158}    &  \textbf{1.9130}     &    34.8674   &   \textbf{1.7040}  & \textbf{37.8432} & \textbf{1.2202} \\
    \hline
    \end{tabular}}%
  \caption{Comparisons on large-scale datasets.}
  \label{tab:larger}%
\end{table}%

\section{Conclusion}
This paper presents a novel Transformer network, i.e., ESSAformer, for the single-hyperspectral image super-resolution task. The Transformer has an iterative refinement architecture to encode the feature representation and context information at different scales. Besides, we utilize the characteristics in HSI and propose a particular ESSA attention mechanism to effectively improve the data efficiency and model performance by involving channel-wise inductive biases. The model also significantly relieves the computation burden with strict theoretical support from kernel machines. Thanks to the particular design, the model builds long-range dependencies and produces better restoration results without involving much compute cost. Extensive experiments on different datasets at various super-resolution scales demonstrate the SOTA performance of ESSAformer regarding both visual quality and objective metrics.

\section{Acknowledgement}
This work was supported in part by the National Natural Science Foundation of China under Grants 62272363, 62036007, 62061047, 62176195, and U21A20514, the Young Elite Scientists Sponsorship Program by CAST under Grant 2021QNRC001, the Youth Talent Promotion Project of Shaanxi University Science and Technology Association under Grant 20200103, the Special Project on Technological Innovation and Application Development under Grant No.cstc2020jscx-dxwtB0032, and the Chongqing Excellent Scientist Project under Grant No. cstc2021ycjh-bgzxm0339. 

\newpage
{\small
\bibliographystyle{ieee_fullname}
\bibliography{egbib}
}

\clearpage
\appendix
\section{Supplementary materials}
\subsection{Ablation study on different attention types}
To study the effectiveness of the proposed method, we use the ESSAformer structure with different attention types, i.e., conventional MHSA, inductive bias-induced SCC attention and the proposed ESSA, for a thoroughly comparison. The results of PSNR, SSIM, and SAM are given in Table \ref{tab:essa ablation}. Different from the test set settings in the paper, we crop the test image from 512 $\times$ 512 to 128 $\times$ 128 due to the huge computational and GPU footprint burden in MHSA and SCC attention. Thanks to the channel-wise inductive bias in SCC attention, it outperforms MHSA significantly on all the metrics. Meanwhile, ESSA has significantly less computation cost and achieves the performance on par with the SCC attention. The results demonstrate the effectiveness of the proposed ESSA.

\begin{table}[htbp]
  \centering
  \small
    \begin{tabular}{cc|c|ccc}
    \toprule
    \multicolumn{2}{c|}{Method} & Dataset & PSNR  & SSIM  & SAM \\
    \hline
    MHSA  & \multirow{3}[0]{*}{$\times$4} & \multirow{3}[0]{*}{Chikusei} & 41.1242 & 0.952 & 2.3693 \\
    SCC   &       &       & 41.3273 & 0.9539 & 2.3127 \\
    ESSA &       &       & 41.4177 & 0.9554 & 2.3096 \\
    \hline
    MHSA  & \multirow{3}[0]{*}{$\times$4} & \multirow{3}[0]{*}{Cave} & 44.5341 & 0.969 & 6.9794 \\
    SCC   &       &       & 44.9210 & 0.9720 & 5.1792 \\
    ESSA  &       &       & 45.2727 & 0.9710 & 4.9596 \\
    \hline
    MHSA  & \multirow{3}[0]{*}{$\times$4} & \multirow{3}[0]{*}{Pavia} & 29.8248 & 0.8351 & 5.6746 \\
    SCC   &       &       & 30.0249 & 0.8410 & 5.5294 \\
    ESSA &       &       & 30.1598 & 0.8468 & 5.5235 \\
    \bottomrule
    \end{tabular}%
  \caption{Comparison on different attention. The experiments are conducted on three datasets with a scale factor 4$\times$.}
  \label{tab:essa ablation}%
\end{table}%

\begin{figure}[htbp]
  \centering
  
    \includegraphics[width=0.95\linewidth]{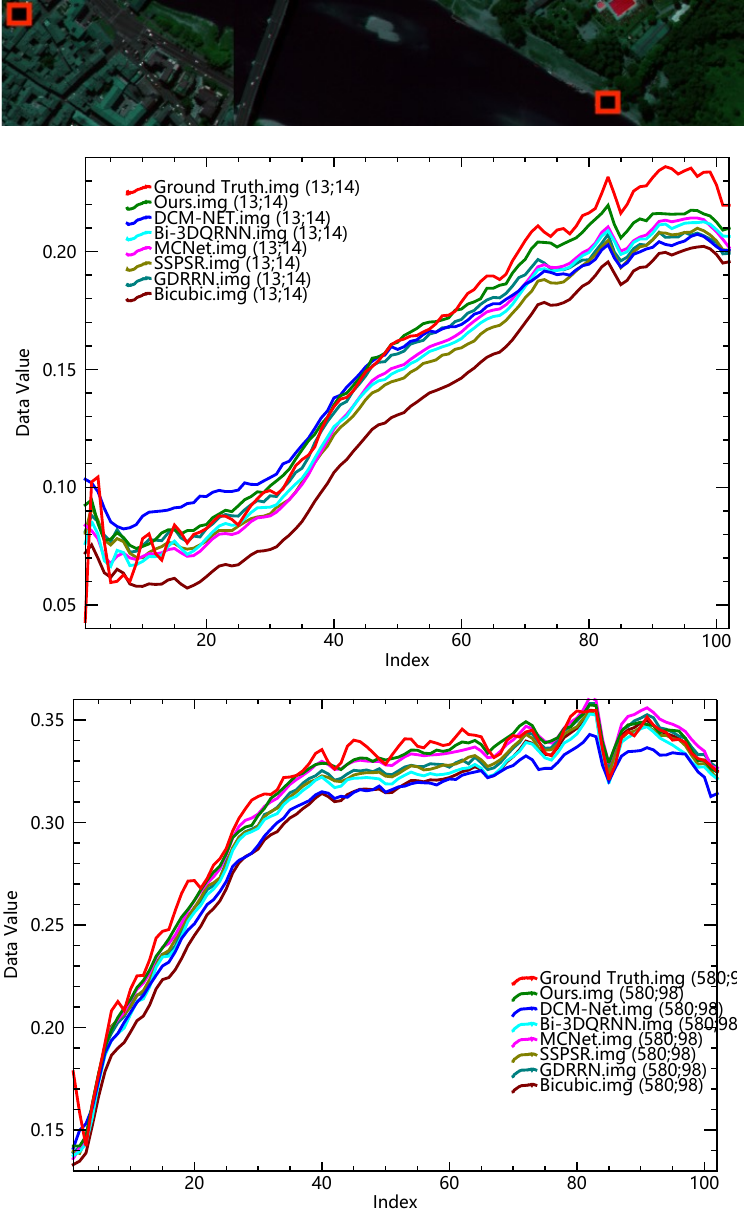}

   \caption{The spectral profiles of a test image from the Pavia dataset.}
   \label{fig:Pavia}
\end{figure}

\begin{figure*}[t]
  \centering
  
    \includegraphics[width=\linewidth]{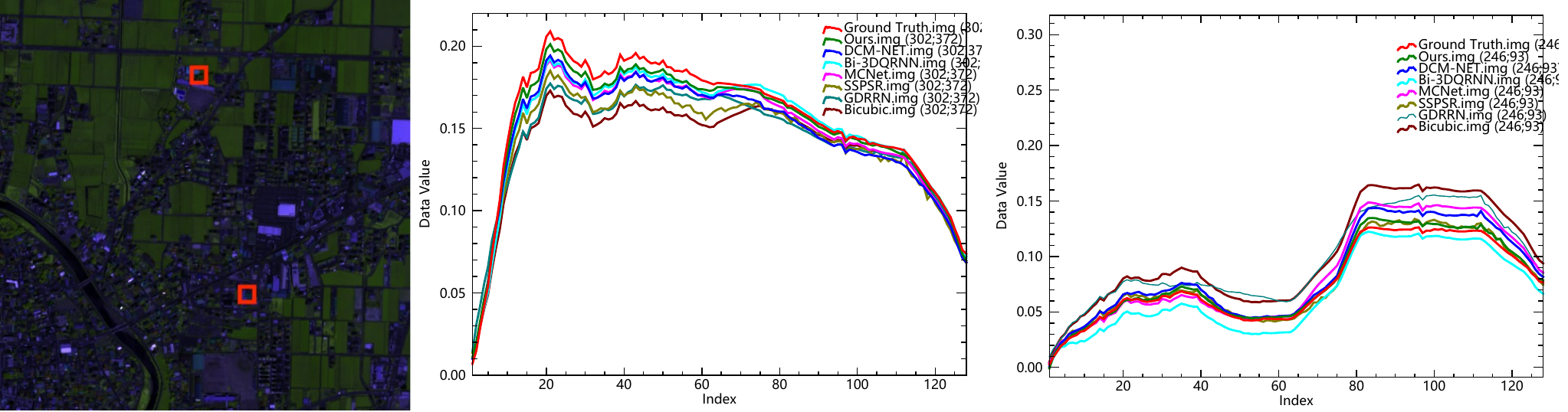}

   \caption{The spectral profiles of a test image from the Chikusei dataset.}
   \label{fig:Chikusei1}
\end{figure*}

\begin{figure*}[t]
  \centering
  
    \includegraphics[width=\linewidth]{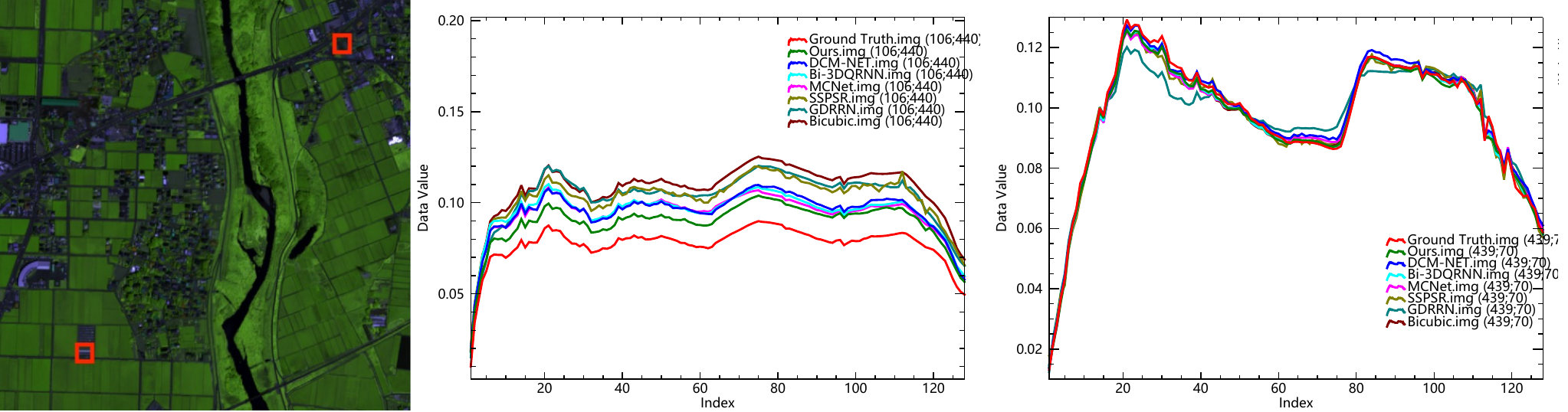}

   \caption{The spectral profiles of a test image from the Chikusei dataset.}
   \label{fig:Chikusei2}
\end{figure*}

\begin{figure*}[t!]
  \centering
  
    \includegraphics[width=\linewidth]{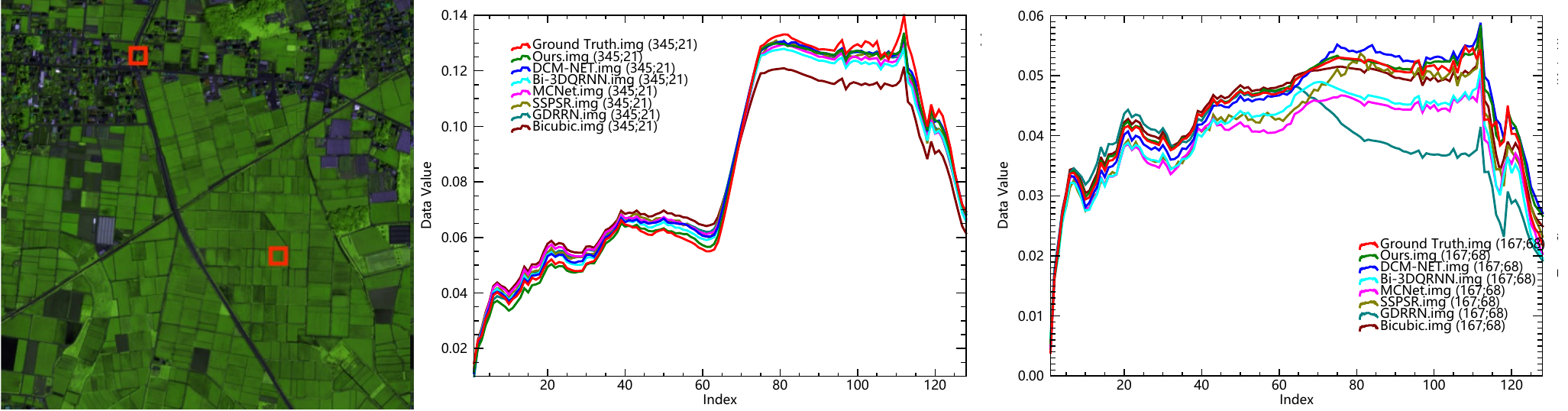}

   \caption{The spectral profiles of a test image from the Chikusei dataset.}
   \label{fig:Chikusei3}
\end{figure*}

\subsection{Qualitative result}

\subsubsection{Spectral profiles}
\begin{figure*}[t!]
  \centering
  
    \includegraphics[width=0.7\linewidth]{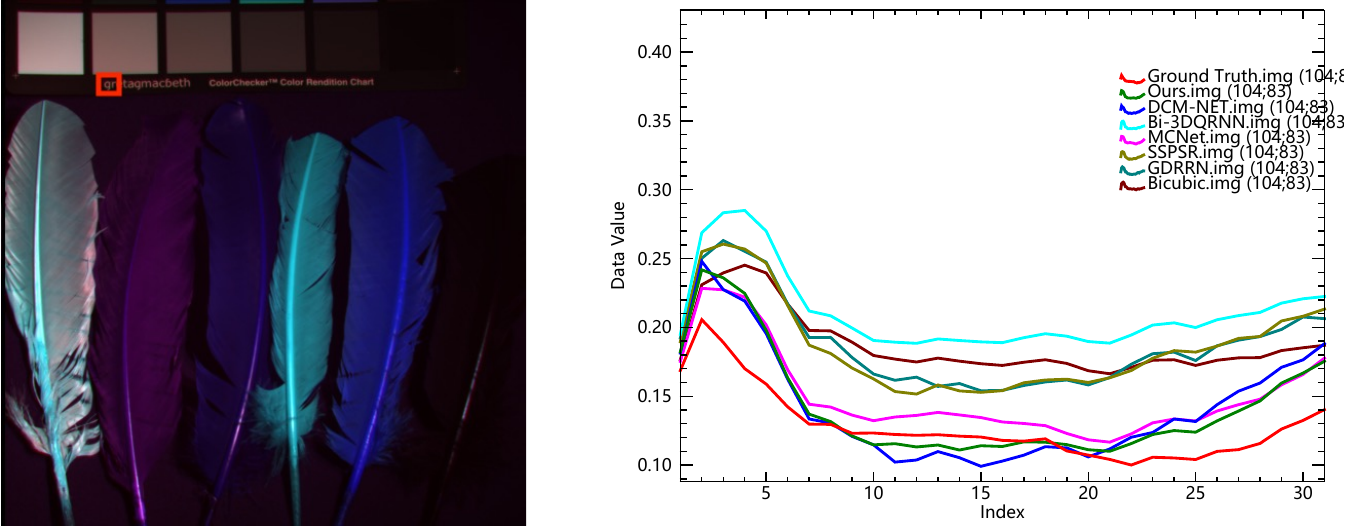}

   \caption{The spectral profiles of a test image from the Cave dataset.}
   \label{fig:Cave1}
\end{figure*}

\begin{figure*}[t!]
  \centering
  
    \includegraphics[width=0.7\linewidth]{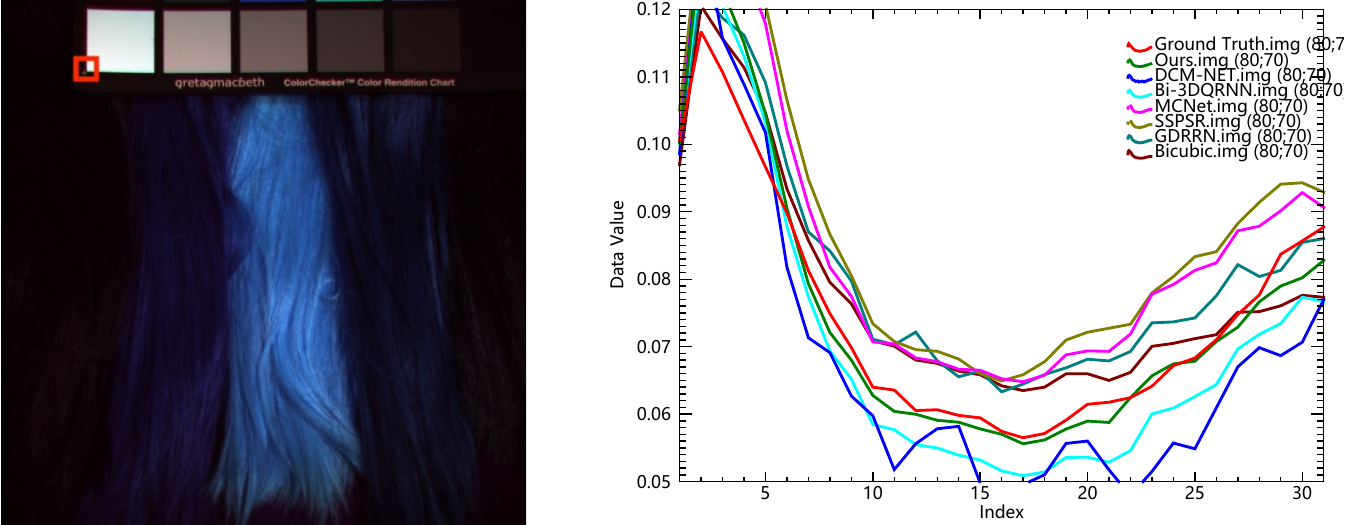}

   \caption{The spectral profiles of a test image from the Cave dataset.}
   \label{fig:Cave4}
\end{figure*}

First, we select several images in the test set from Pavia, Chikusei, and Cave and plot the spectral profiles of the red circle regions as shown in Figure~\ref{fig:Pavia}, \ref{fig:Chikusei1}, \ref{fig:Chikusei2}, \ref{fig:Cave1}, \ref{fig:Cave4}, and \ref{fig:Chikusei3}, respectively. It can be seen that our method, i.e., the green line, recovers the most information and is the closest to the groundtruth red line in all figures at all data values. For example, the green line approaches the red most in the high-value range, i.e., the index between 60 and 100, in Figure~\ref{fig:Pavia}, while others fail to reach the peaks compared to ESSAformer. In contrast, as shown in Figure~\ref{fig:Chikusei2}, all the models tend to generate higher values in spectral profiles compared to the groundtruth. Our ESSAformer, however, relieves this tendency most thanks to the introduced channel-wise inductive bias in ESSA attention.

\subsubsection{Visualized absolute error}
\begin{figure*}[h]
  \centering
    \includegraphics[width=\linewidth]{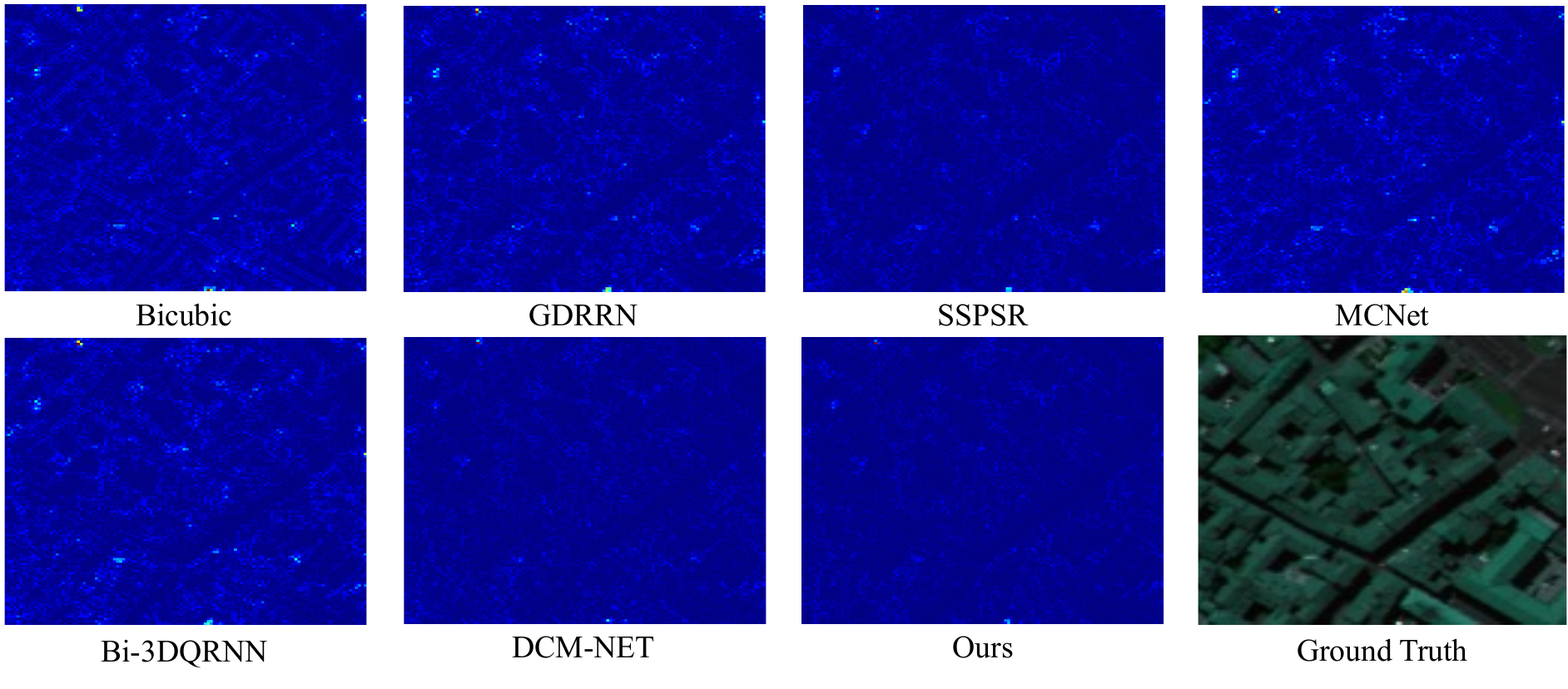}
   \caption{The absolute error map of a test image from the Pavia dataset.}
   \label{fig:Pavia absolute}
   \includegraphics[width=\linewidth]{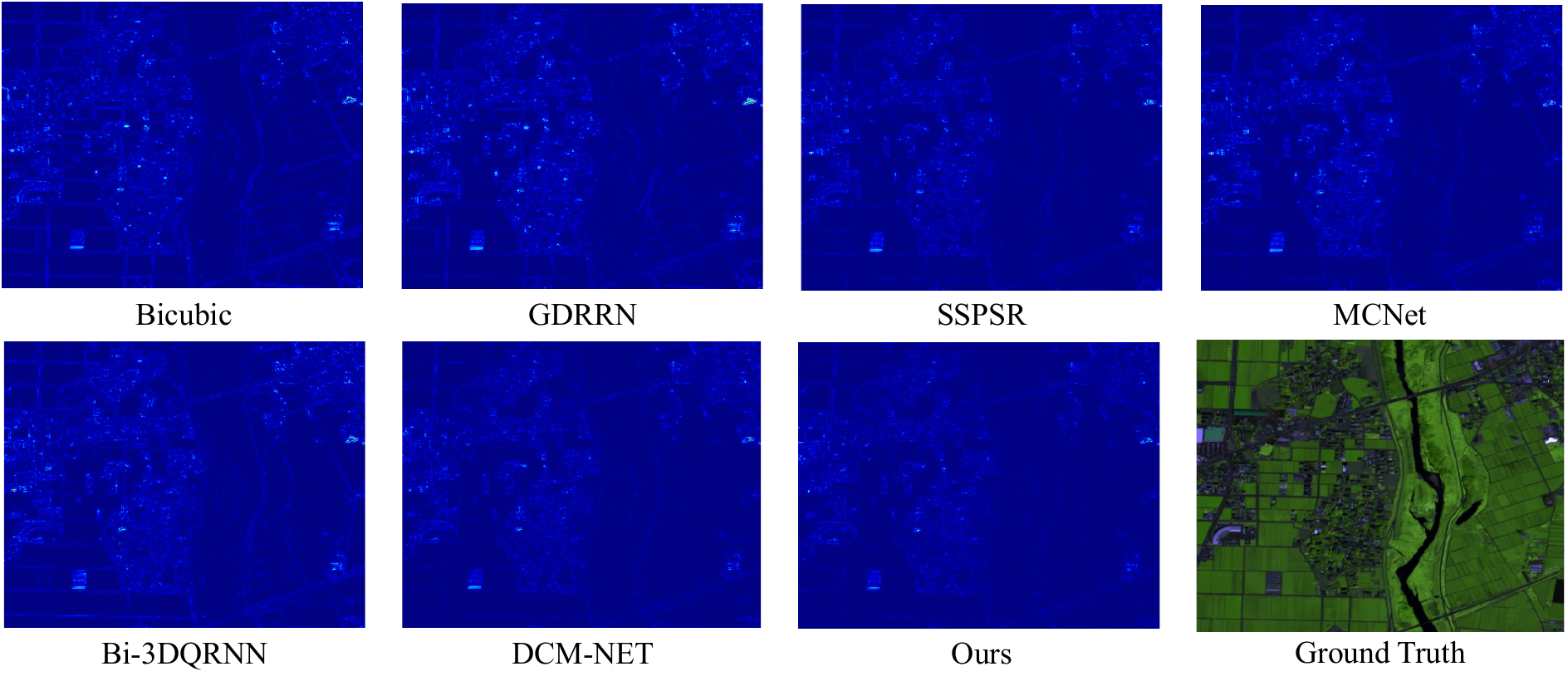}

   \caption{The absolute error map of a test image from the Chikusei dataset.}
   \label{fig:Chikusei absolute1}
\end{figure*}

\begin{figure*}[h]
  \centering
  \setlength{\belowcaptionskip}{-1mm}
    \includegraphics[width=\linewidth]{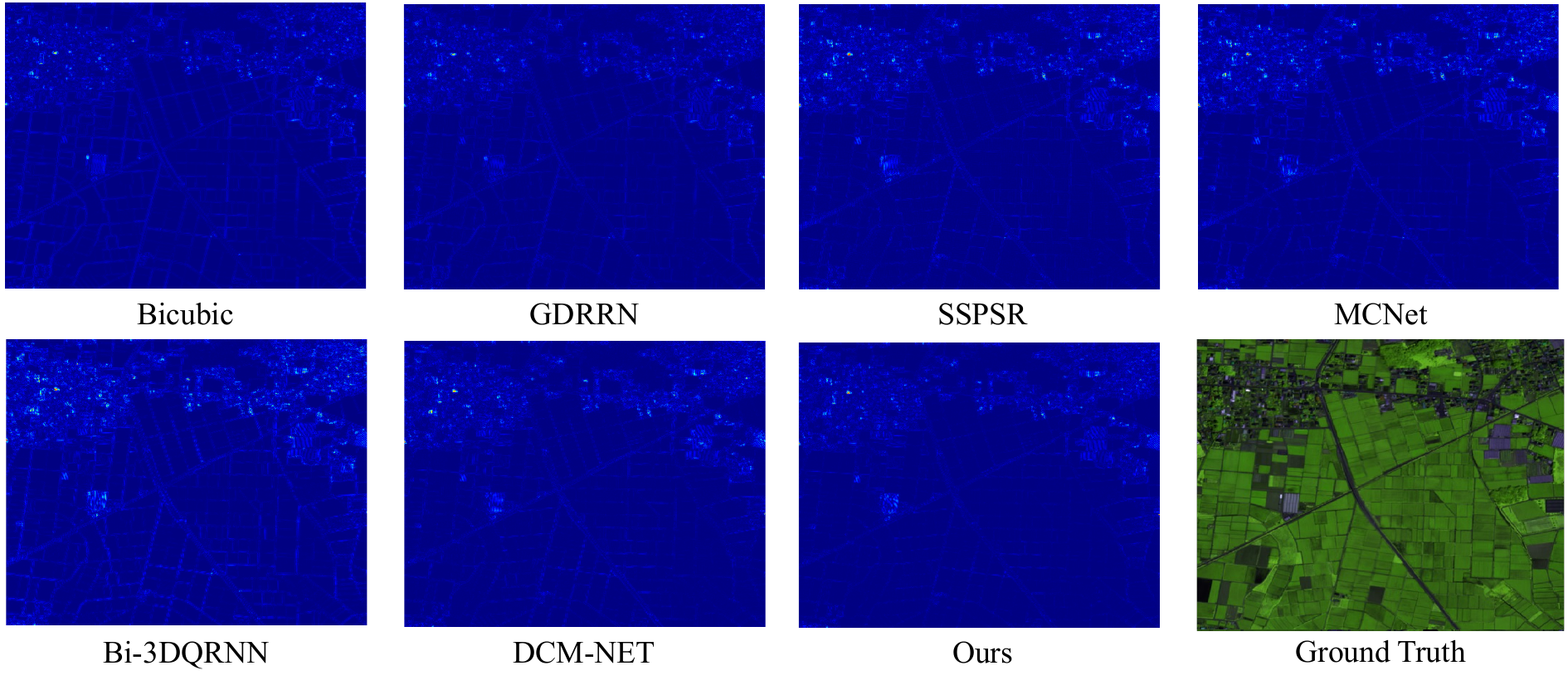}

   \caption{The absolute error map of a test image from the Chikusei dataset.}
   \label{fig:Chikusei absolute3}

    \centering
    \setlength{\belowcaptionskip}{-1cm}
    \includegraphics[width=\linewidth]{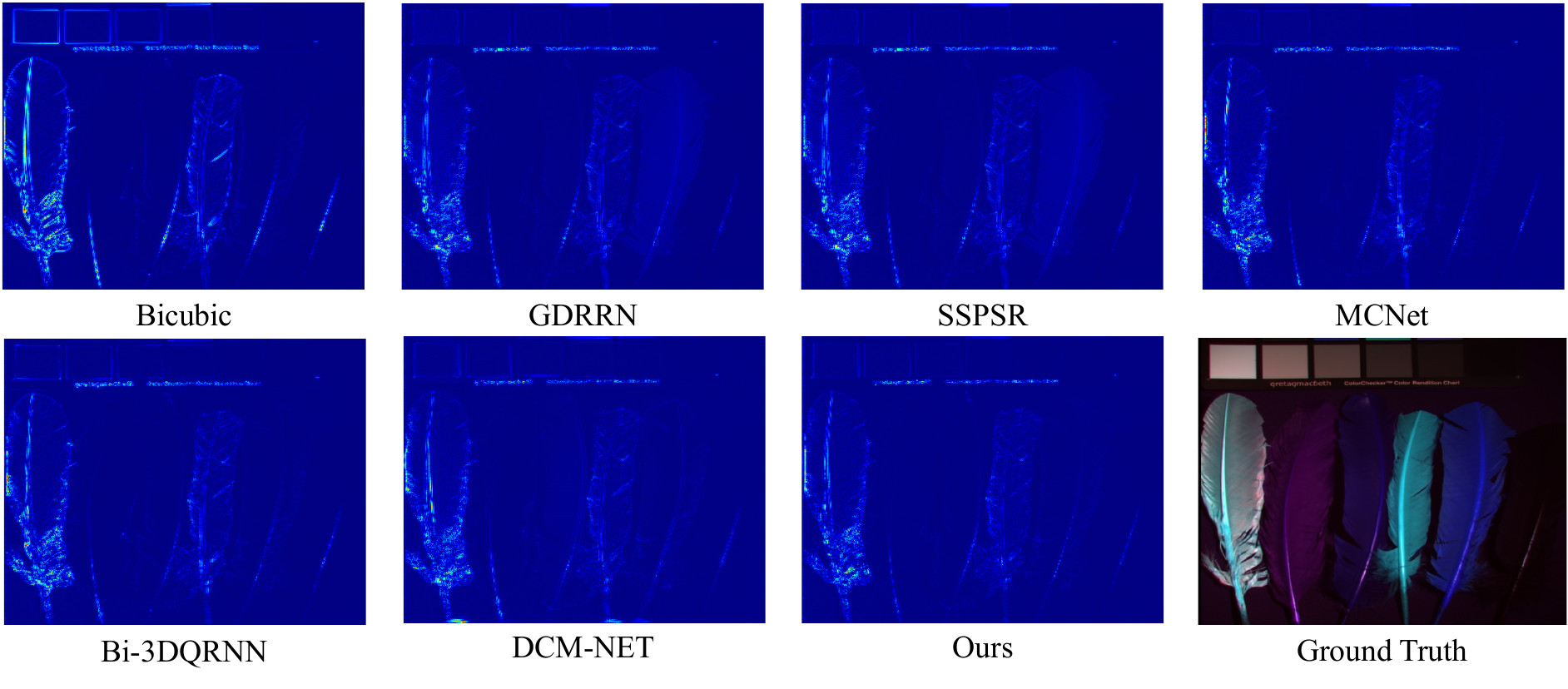}

   \caption{The absolute error map of a test image from the Cave dataset.}
   \label{fig:Cave absolute2} 
     
     \centering
 
    \includegraphics[width=\linewidth]{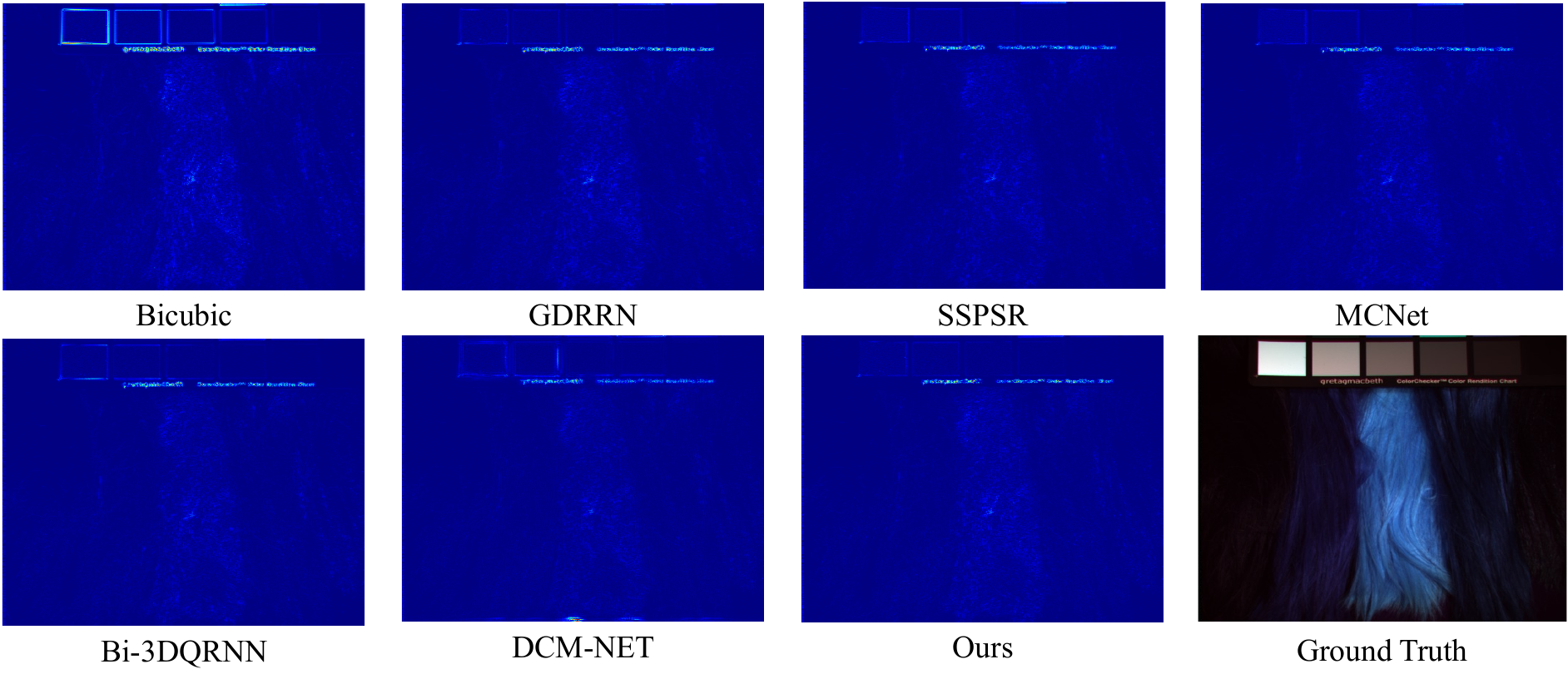}

   \caption{The absolute error map of a test image from the Cave dataset.}
   \label{fig:Cave absolute3}
\end{figure*}

We visualize the absolute error maps of different methods as shown in Figure~ \ref{fig:Pavia absolute}, \ref{fig:Chikusei absolute1}, \ref{fig:Chikusei absolute3}, \ref{fig:Cave absolute2}, and \ref{fig:Cave absolute3}. The original images after re-formatting to RGB ones are also given in each figure for reference. The pixels with dark colors denote the small error and the bright refer to having a large absolute error. From the figure, we can see that ESSAformer generates the images with the least textures and thus obtains the best performance. For example, our method produces results with the slightest difference from the ground truth, better recovering the bird's eye view image of Pavia city, as shown in Figure~\ref{fig:Pavia absolute}. Besides, the residual maps of other methods in Figure \ref{fig:Chikusei absolute1} and \ref{fig:Chikusei absolute3} show the `road' clearly while ESSAformer has a strong ability to restore such edges in its outputs. When comparing the living area with abundant variance in the figures, ESSAformer also has darker results than the others, demonstrating its superiority for processing such challenging regions. Besides, as shown in the bottom leather and upper `rectangle' regions in Figure \ref{fig:Cave absolute2}, ESSAformer effectively restores the details, leading to a conclusion similar to the previous analysis.

\end{document}